\documentclass[11pt]{article}
\usepackage{graphicx}
\usepackage{multirow}
\usepackage{amsmath,amssymb,amsfonts}
\usepackage{amsthm}
\usepackage[title]{appendix}
\usepackage{textcomp}
\usepackage{manyfoot}
\usepackage{booktabs}

\usepackage{adjustbox}   
\usepackage{makecell}
\usepackage{array}
\usepackage{siunitx}
\usepackage{pifont}
\usepackage{placeins}
\usepackage{subcaption}

\usepackage{algorithm}
\usepackage{algorithmicx}
\usepackage{algpseudocode}
\usepackage{listings}\usepackage[super]{cite}
\usepackage{authblk}
\usepackage{mathpazo}
\usepackage{siunitx}\usepackage{amsmath,amssymb,amsfonts}
\usepackage{graphicx}
\usepackage{textcomp}
\usepackage{array}
\usepackage{multirow}
\usepackage{lineno}
\usepackage{hyperref}
\usepackage[table,xcdraw]{xcolor}
\usepackage{algpseudocode}
\usepackage{algorithm}
\usepackage{subcaption}
\usepackage{pifont}

\usepackage[top=1in, bottom=1in, left=0.75in, right=0.75in]{geometry}
\def\BibTeX{{\rm B\kern-.05em{\sc i\kern-.025em b}\kern-.08em
    T\kern-.1667em\lower.7ex\hbox{E}\kern-.125emX}}
\title{OCT-FedSIR: Toward Trustworthy Federated Ophthalmic Learning under Annotation Noise}

\author[1]{Sina Gholami}
\author[1]{Abdulmoneam Ali}
\author[1]{Tania Haghighi}
\author[1]{Rashadul H. Badhon}
\author[1]{Behafarin Emam}
\author[2]{Sally S. Y. Ong}
\author[2]{Atalie C. Thompson}
\author[4]{Theodore Leng}
\author[1]{Ahmed Arafa}
\author[3]{Jennifer I. Lim}
\author[1]{Minhaj Nur Alam\thanks{Corresponding author: malam8@charlotte.edu}}

\affil[1]{Department of Electrical and Computer Engineering, University of North Carolina at Charlotte, Charlotte, NC, USA}
\affil[2]{Department of Ophthalmology, Wake Forest School of Medicine, Winston-Salem, NC, USA}
\affil[3]{Department of Ophthalmology and Visual Sciences, University of Illinois Chicago, Chicago, IL, USA}
\affil[4]{Byers Eye Institute at Stanford, Stanford University School of Medicine, Stanford, CA, USA}
\date{}
\begin{document}
\maketitle

\begin{abstract}
Federated learning enables collaborative model development without centralizing patient data, but the reliability of annotations retained at participating institutions cannot always be assumed. This creates a particular challenge in ophthalmic imaging, where legitimate differences in disease prevalence and local class composition may resemble changes produced by corrupted supervision. We introduce OCT-FedSIR, a reliability-aware spectral framework for federated OCT classification under client-dependent annotation noise and heterogeneous data distributions. OCT-FedSIR combines class-balanced spectral estimation, Stage-I logit adjustment, and complementary spectral descriptors to distinguish annotation-related disruption from statistical heterogeneity, followed by selective spectral relabeling and noise-aware federated optimization. The framework was evaluated on three ophthalmic classification tasks using the Kermany, University of Illinois Chicago, and Wake Forest datasets under symmetric and structured asymmetric annotation noise and three levels of non-IID heterogeneity. Across 117 experimental conditions, OCT-FedSIR achieved a mean accuracy of 86.73\%, compared with 79.94\% for RoFL and 78.75\% for FedCorr. Under the controlled corruption settings, OCT-FedSIR correctly separated clients with original and corrupted annotations across all evaluated conditions, whereas the original FedSIR identification procedure showed reduced performance, particularly under asymmetric noise. Spectral relabeling recovered 77.2\% of intentionally corrupted annotations on average, with a mean correction precision of 91.3\% and a mean false-correction rate of 3.5\%. Retaining corrected clients also outperformed spectral pruning by 9.30 percentage points on average. These findings identify annotation reliability as an important consideration for trustworthy ophthalmic federated learning and show that corrupted supervision can often be corrected without discarding potentially informative client data.
\end{abstract}


\section*{Introduction}
Optical coherence tomography (OCT) has become a cornerstone of retinal care by enabling detailed visualization of retinal morphology that supports disease detection, characterization, and longitudinal monitoring. Deep learning has shown promise for automated OCT interpretation, including disease classification from retinal B-scans, detection and quantification of pathological features, and clinically oriented diagnosis and referral from volumetric OCT imaging \cite{lee2017deep,kermany,schlegl2018,Fauw}. Together, these studies, including our recent studies \cite{haghighi2025compact,siraz2025multi,jannat2026multi}, have established that diagnostically relevant information can be learned directly from OCT and have motivated the development of models capable of operating across broader and more heterogeneous clinical settings.

Translation across institutions, however, requires models to accommodate differences in patient populations, disease prevalence, imaging systems, acquisition protocols, and clinical workflows. Because ophthalmic data are naturally distributed across hospitals and imaging centers, centralizing sufficiently diverse datasets may be constrained by patient-privacy requirements, including compliance with the Health Insurance Portability and Accountability Act (HIPAA), as well as data-ownership and institutional-governance considerations. Federated learning (FL) \cite{macmahan} provides an alternative by enabling multiple institutions to optimize a shared model while retaining patient data locally. In ophthalmology, FL has been investigated across retinal image segmentation, diabetic retinopathy (DR) classification, retinopathy of prematurity, OCT-based age-related macular degeneration (AMD) classification, and glaucoma detection \cite{lo2021federated,lu2022federated,gholami2023federated,ran2024developing}. More recent studies have extended federated and distributed learning to multi-disease retinal classification, self-supervised representation learning, masked-autoencoder pretraining, and personalized OCT classification \cite{nabil2025federated,lau2025fedted,gholami-distributed,fedsim}. Collectively, these studies demonstrate the feasibility of collaborative ophthalmic model development without centralizing raw images, but have largely focused on statistical and domain heterogeneity rather than whether the supervision available at participating clients is itself reliable.

This distinction is important for trustworthy FL. Retaining patient images locally addresses where data are stored, but does not ensure that the annotations used to train a shared model are reliable. Retinal labels may be assigned prospectively by specialist graders or derived retrospectively from clinical documentation, diagnostic codes, and electronic health records, and these sources do not provide uniform annotation quality \cite{yonamine2024comparison}. Agreement may vary with grader expertise, diagnostic criteria, image quality, disease severity, coexisting pathology, and the availability of complementary clinical information. Intergrader variability has been documented throughout medical and ophthalmic image interpretation, particularly for conditions represented along a severity continuum \cite{ju2022improving,campbell2022artificial}. OCT presents an additional challenge because diagnostically distinct retinal disorders may share overlapping structural findings, and grading decisions may become uncertain near disease or severity boundaries. Annotation errors may therefore be structured rather than uniformly distributed across samples or diagnostic categories.

The consequences of unreliable supervision can be substantial. Label noise is known to impair generalization in medical image analysis \cite{karimi2020noisy,shi2024labelnoise}. In OCT classification, compensating for label-noise rates of 10\%, 15\%, and 20\% has been estimated to require approximately four-, nine-, and fourteen-fold increases in training data, respectively, to recover performance obtained with clean annotations \cite{miladinovic2024evaluating}. Retinal imaging studies have similarly demonstrated progressive performance degradation with increasing annotation noise and partial recovery following automated label cleaning \cite{lin2025efficiency}. Correction itself, however, introduces another risk: corrupted annotations may remain undetected, while labels that were originally correct may be modified incorrectly \cite{lin2025efficiency}. Reliable learning under annotation noise therefore requires more than improving downstream accuracy; it requires determining which supervision is unreliable, whether corrupted labels can be recovered, and whether correction can be performed without substantially damaging annotations that were already correct.

This problem becomes particularly challenging in FL because annotation reliability must be inferred from heterogeneous local datasets without centralizing the underlying images. Existing noisy-label FL methods have addressed unreliable supervision through reliable-sample selection, confidence- or loss-based weighting, noisy-client identification, knowledge distillation (KD), robust aggregation, and optimization of corrected label distributions \cite{rofl,fbnll,rhfl,fedcorr,fednoro,fedned,fedelc}. FedSIR~\cite{fedsir} introduced a complementary perspective by exploiting the spectral structure of class-specific feature representations. Rather than relying primarily on prediction confidence or training loss, FedSIR uses changes in feature geometry to identify clients affected by label corruption and constructs spectral references from identified clean clients to guide selective relabeling. This representation-level strategy demonstrated strong robustness to severe label noise and non-IID heterogeneity on federated benchmark datasets.

Its translation to ophthalmic data, however, exposes a fundamental challenge: \emph{an unusual client is not necessarily an unreliable client}. Differences in local disease prevalence, class imbalance, incomplete representation of diagnostic categories, and strongly non-IID client distributions can alter learned feature geometry even when annotations are entirely correct. Consequently, spectral deviations caused by legitimate differences in clinical case mix may resemble those caused by annotation corruption. This creates an important challenge for trustworthy ophthalmic FL: distinguishing genuine clinical heterogeneity from unreliable supervision before using that distinction to modify labels or client contributions.

To address this problem, we introduce \textbf{OCT-FedSIR}, a spectral-guided framework for federated OCT classification under client-dependent annotation noise and heterogeneous local data distributions. OCT-FedSIR strengthens client characterization through class-balanced spectral estimation, Stage-I logit adjustment (LA), and complementary similarity descriptors designed to reduce sensitivity to differences in local class composition. Rather than automatically excluding clients identified as unreliable, spectral information learned from reliable clients is used to guide selective relabeling, allowing potentially informative local data to remain available for federated optimization. Importantly, the framework evaluates annotation recovery itself in addition to downstream classification performance, enabling both successful recovery and harmful correction of originally correct labels to be quantified. The main contributions of this study are: 
\begin{itemize} 
\item We introduce \textbf{OCT-FedSIR}, a reliability-aware spectral framework for ophthalmic FL under annotation noise. OCT-FedSIR combines class-balanced spectral estimation, Stage-I LA, and complementary spectral similarity descriptors to improve discrimination between annotation-related disruption and legitimate variation in local disease composition. 
\item We evaluate spectral relabeling at the individual-sample level by quantifying recovery of corrupted annotations, preservation of originally correct labels, correction precision, and residual label-noise rate. We further compare correction and retention of clients identified as unreliable with spectral pruning to determine whether useful local information can be recovered rather than discarded.
\item We evaluate OCT-FedSIR across three distinct ophthalmic classification settings: multiclass OCT disease classification using the Kermany dataset, ordinal DR classification using a University of Illinois Chicago (UIC) cohort, and binary geographic atrophy (GA) classification using a Wake Forest (WF) cohort. Experiments span symmetric and structured asymmetric annotation noise, multiple noise ratios, and varying degrees of non-IID client heterogeneity, with comparisons against conventional FL and dedicated noisy-label FL approaches.
\end{itemize}
Through these experiments, we investigate whether spectral characteristics can distinguish annotation-related disruption from legitimate client heterogeneity, how reliably corrupted supervision can be recovered without damaging originally correct annotations, and whether retaining corrected clients provides greater benefit than excluding them from federated optimization. More broadly, this study examines annotation reliability as an important and previously underexplored requirement for trustworthy ophthalmic FL.

\begin{figure}[ht]
    \centering
    \includegraphics[width=0.95\textwidth]{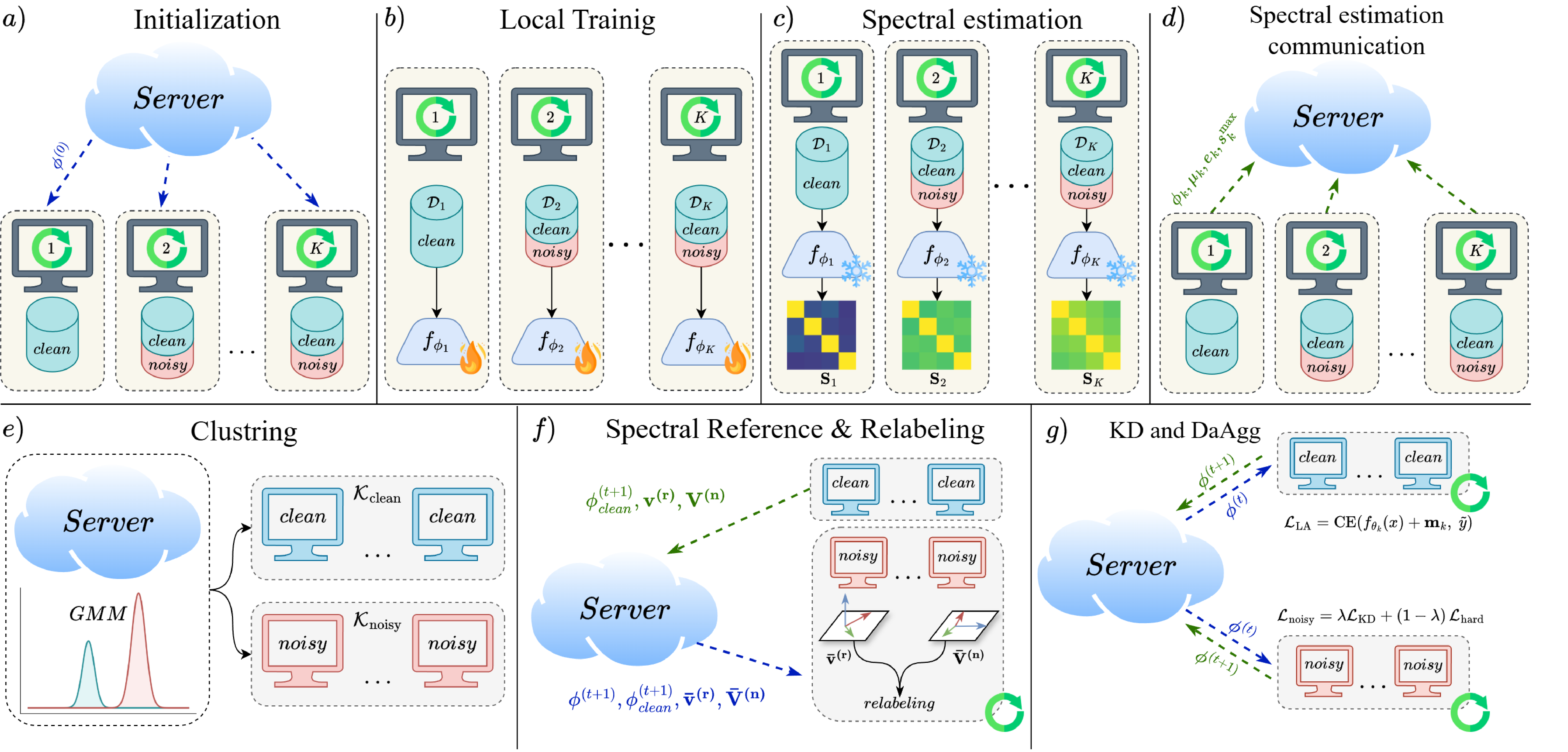}
    \caption{Overview of the OCT-FedSIR framework.
    \textbf{a}, Global initialization is distributed to participating clients.
    \textbf{b}, Local training.
    \textbf{c--e}, Spectral statistics are estimated, communicated to the server,
    and used to identify clean and noisy clients.
    \textbf{f}, Clean-client spectral information is used to construct
    representative and residual references for relabeling.
    \textbf{g}, Noise-aware federated optimization using LA, KD, and DaAgg.}
    \label{fig:stages}
\end{figure}

\section*{Results}
We evaluated OCT-FedSIR across three distinct ophthalmic classification tasks: multiclass OCT disease classification using Kermany, ordinal DR severity classification using UIC, and binary GA classification using WF. Experiments included controlled symmetric and structured asymmetric annotation noise across multiple noise ratios and three degrees of client heterogeneity, $\alpha\in\{2.0,0.5,0.1\}$, with lower values representing increasingly heterogeneous non-IID client distributions. OCT-FedSIR was compared with conventional FL baselines, including FedAvg~\cite{macmahan} and FedProx~\cite{fedprox}, and dedicated noisy-label FL methods, including RoFL~\cite{rofl}, RHFL~\cite{rhfl}, FedLSR~\cite{fedlsr}, FedCorr~\cite{fedcorr}, FedNed~\cite{fedned}, FedELC~\cite{fedelc}, and FedNoRo~\cite{fednoro}. We additionally evaluated a pruning baseline in which clients identified as having corrupted annotations were excluded from subsequent federated optimization, and an all-clean FedAvg reference in which all participating clients retained their original annotations.

\subsection*{Overall performance}
Table~\ref{tab:overall_summary} summarizes classification accuracy across
datasets, noise mechanisms, noise levels, and client-heterogeneity
settings. For each dataset, symmetric results were averaged across 21
experimental conditions, corresponding to seven noise levels ranging from $30\%$ to $90\%$ with $10\%$ increments and three
values of $\alpha$, whereas asymmetric results were averaged across 18
conditions, corresponding to six noise levels ranging from 40\% to 90\% and the same three values
of $\alpha$. The overall result therefore summarizes 117 condition-level mean
accuracies.

\begin{table*}[t]
\centering
\caption{
Mean classification accuracy (\%) across label-noise levels and
client-heterogeneity settings. Symmetric and asymmetric results are averaged
over 21 and 18 experimental conditions per dataset, respectively, and overall
accuracy is averaged across all 117 conditions. Bold and underline indicate
the best- and second-best-performing noisy-label FL methods, respectively.
Pruning and all-clean FedAvg are shown as reference configurations.
}
\label{tab:overall_summary}

\begin{adjustbox}{width=\textwidth}
\begin{tabular}{lccccccc}
\toprule
& \multicolumn{2}{c}{Multi-class (Kermany)}
& \multicolumn{2}{c}{Ordinal DR severity (UIC)}
& \multicolumn{2}{c}{Binary GA (WF)}
& \multirow{2}{*}{Overall} \\
\cmidrule(lr){2-3}
\cmidrule(lr){4-5}
\cmidrule(lr){6-7}

Method
& Sym. & Asym.
& Sym. & Asym.
& Sym. & Asym.
& \\

\midrule

FedAvg \cite{macmahan}
& 70.13 & 53.79
& 68.71 & 57.10
& 65.82 & 66.44
& 64.02 \\

FedProx \cite{fedprox}
& 69.85 & 52.28
& 67.40 & 55.80
& 64.89 & 66.32
& 63.11 \\

RoFL \cite{rofl}
& \underline{87.56} & \underline{82.24}
& \underline{81.69} & 81.94
& \underline{72.45} & \underline{73.49}
& \underline{79.94} \\

RHFL \cite{rhfl}
& 44.53 & 33.53
& 52.65 & 38.39
& 59.64 & 45.42
& 46.20 \\

FedLSR \cite{fedlsr}
& 67.46 & 62.32
& 71.55 & 64.99
& 69.81 & 71.86
& 68.12 \\

FedCorr \cite{fedcorr}
& 86.19 & 81.76
& 81.57 & \underline{83.43}
& 68.88 & 70.62
& 78.75 \\

FedNed \cite{fedned}
& 69.21 & 54.20
& 67.46 & 56.77
& 64.72 & 59.52
& 62.38 \\

FedELC \cite{fedelc}
& 60.01 & 50.84
& 64.36 & 55.79
& 61.56 & 62.36
& 59.37 \\

FedNoRo \cite{fednoro}
& 64.23 & 61.48
& 76.07 & 68.04
& 70.24 & 69.62
& 68.43 \\

\textbf{OCT-FedSIR (Ours)}
& \textbf{94.76} & \textbf{93.59}
& \textbf{88.39} & \textbf{89.57}
& \textbf{75.79} & \textbf{78.47}
& \textbf{86.73} \\

\midrule

Pruning
& 81.23 & 84.20
& 83.47 & 79.97
& 67.22 & 68.53
& 77.43 \\

FedAvg (all clients clean)
& 98.97 & 98.97
& 94.47 & 94.47
& 81.86 & 81.86
& 91.76 \\

\bottomrule
\end{tabular}
\end{adjustbox}
\end{table*}

Across all 117 evaluated conditions, OCT-FedSIR achieved the highest mean accuracy among methods trained in the presence of noisy clients, reaching 86.73\%. The next strongest methods were RoFL at 79.94\% and FedCorr at 78.75\%,
corresponding to absolute differences of 6.79 and 7.98 percentage points, respectively. 

\begin{figure}[!p]
    \centering


    \begin{subfigure}[t]{0.95\textwidth}
        \centering
        \includegraphics[
            width=\linewidth
        ]{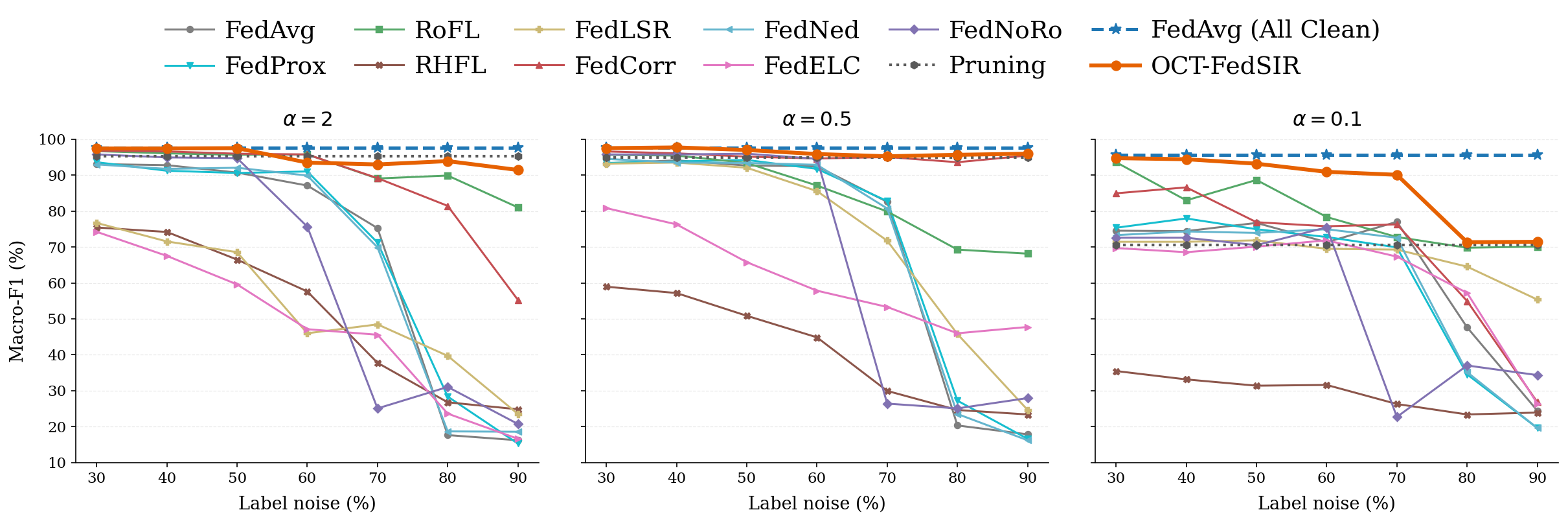}
        \caption{
        Kermany.
        }
        \label{fig:f1_kermany_sym}
    \end{subfigure}

    \vspace{0.5em}


    \begin{subfigure}[t]{0.95\textwidth}
        \centering
        \includegraphics[
            width=\linewidth
        ]{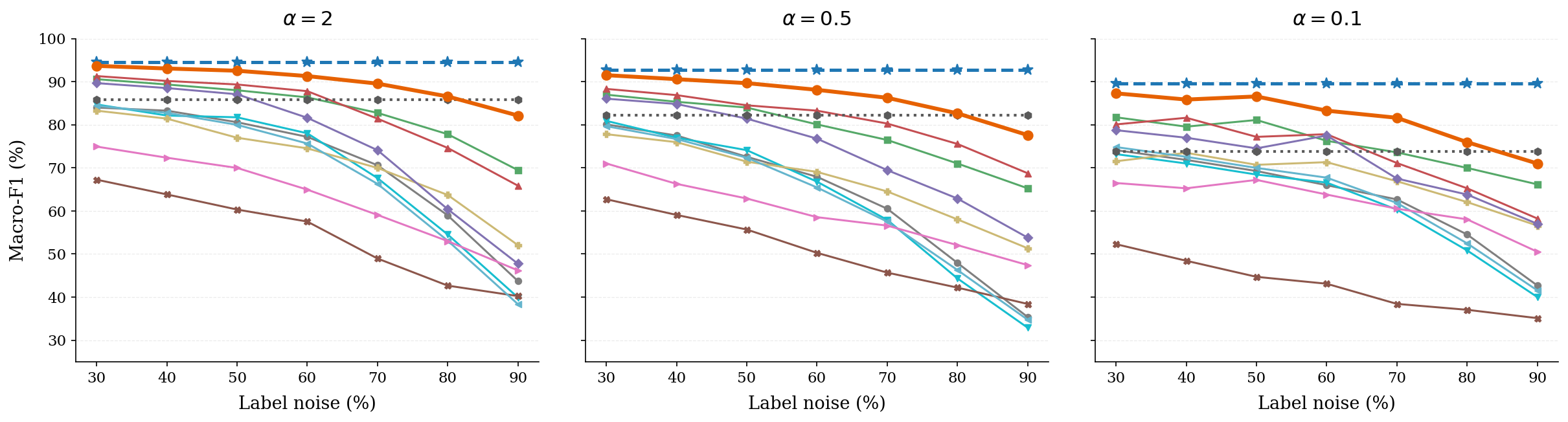}
        \caption{
        UIC.
        }
        \label{fig:f1_uic_sym}
    \end{subfigure}

    \vspace{0.5em}


    \begin{subfigure}[t]{0.95\textwidth}
        \centering
        \includegraphics[
            width=\linewidth
        ]{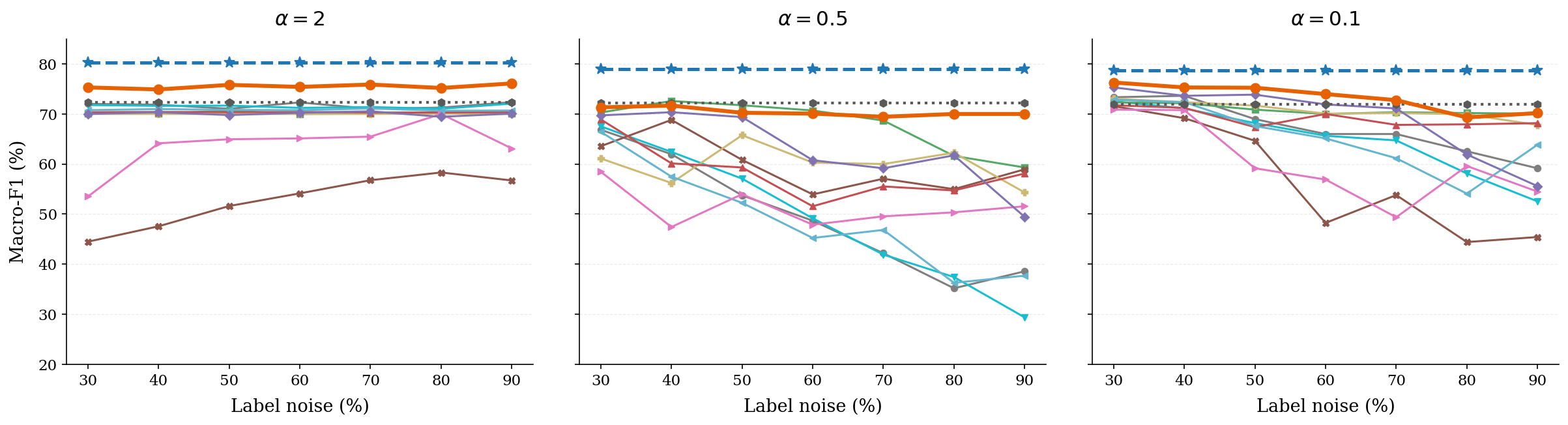}
        \caption{
        WF.
        }
        \label{fig:f1_wf_sym}
    \end{subfigure}


    \caption{
    Macro-F1 performance under symmetric label noise.
    Results are shown across $\alpha \in \{2.0,0.5,0.1\}$ for
    \textbf{a}, Kermany OCT classification;
    \textbf{b}, UIC DR classification; and
    \textbf{c}, WF GA classification.
    The dashed line denotes the all-clean FedAvg reference.
    }
    \label{fig:macro_f1_symmetric}

\end{figure}

\begin{figure}[!p]
    \centering


    \begin{subfigure}[t]{0.95\textwidth}
        \centering
        \includegraphics[
            width=\linewidth
        ]{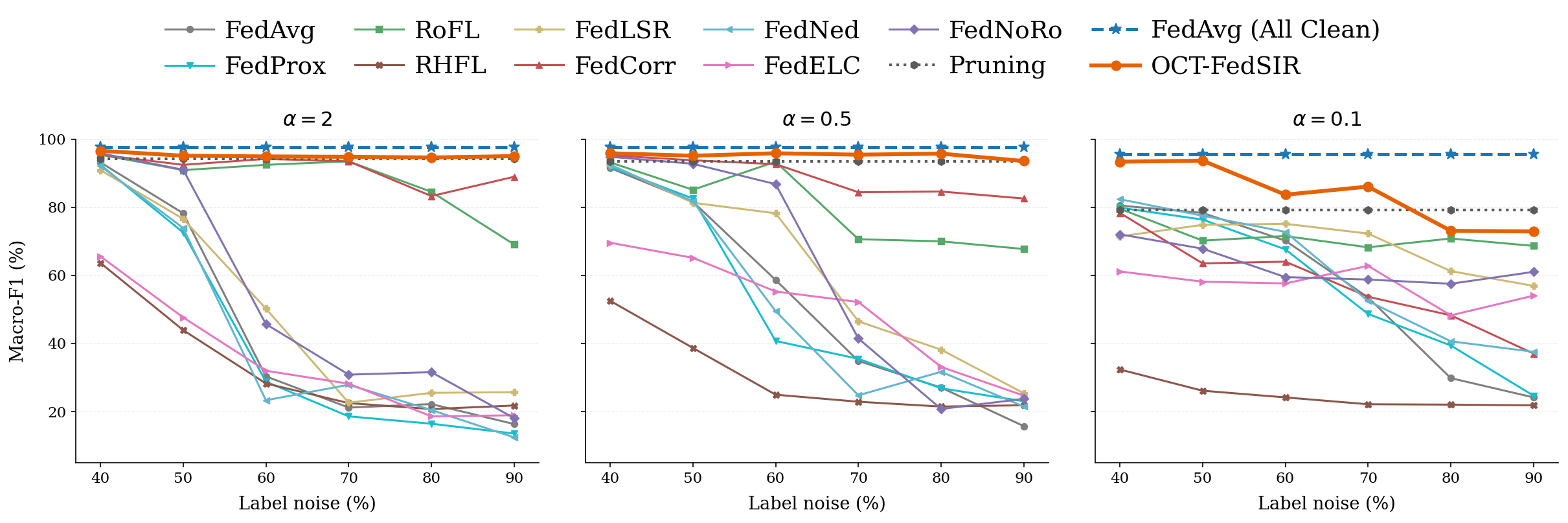}
        \caption{
        Kermany.
        }
        \label{fig:f1_kermany_asym}
    \end{subfigure}

    \vspace{0.5em}


    \begin{subfigure}[t]{0.95\textwidth}
        \centering
        \includegraphics[
            width=\linewidth
        ]{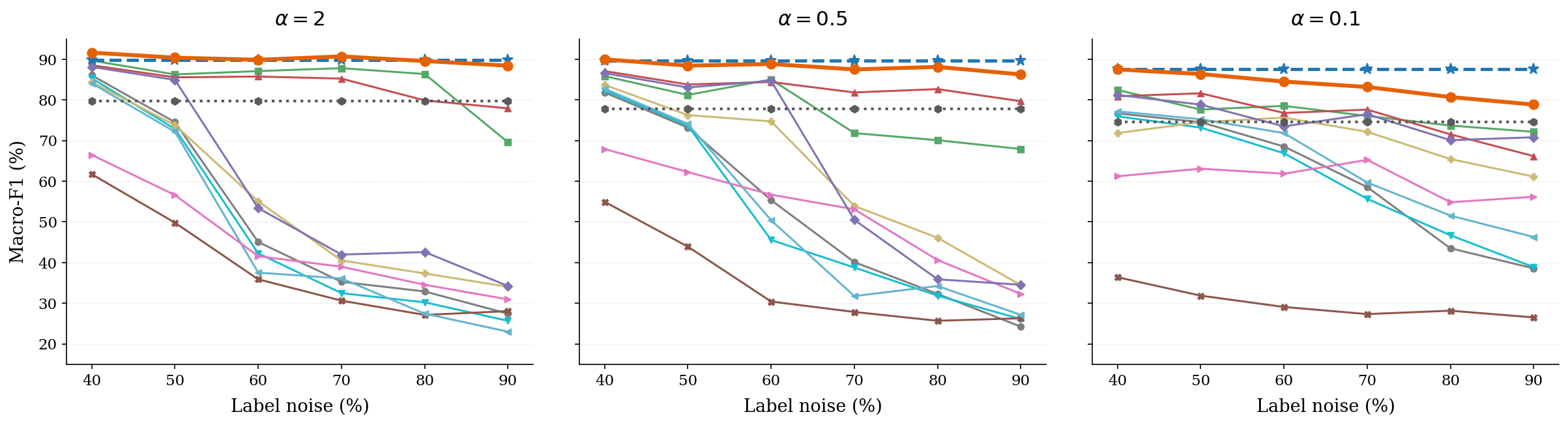}
        \caption{
        UIC.
        }
        \label{fig:f1_uic_asym}
    \end{subfigure}

    \vspace{0.5em}


    \begin{subfigure}[t]{0.95\textwidth}
        \centering
        \includegraphics[
            width=\linewidth
        ]{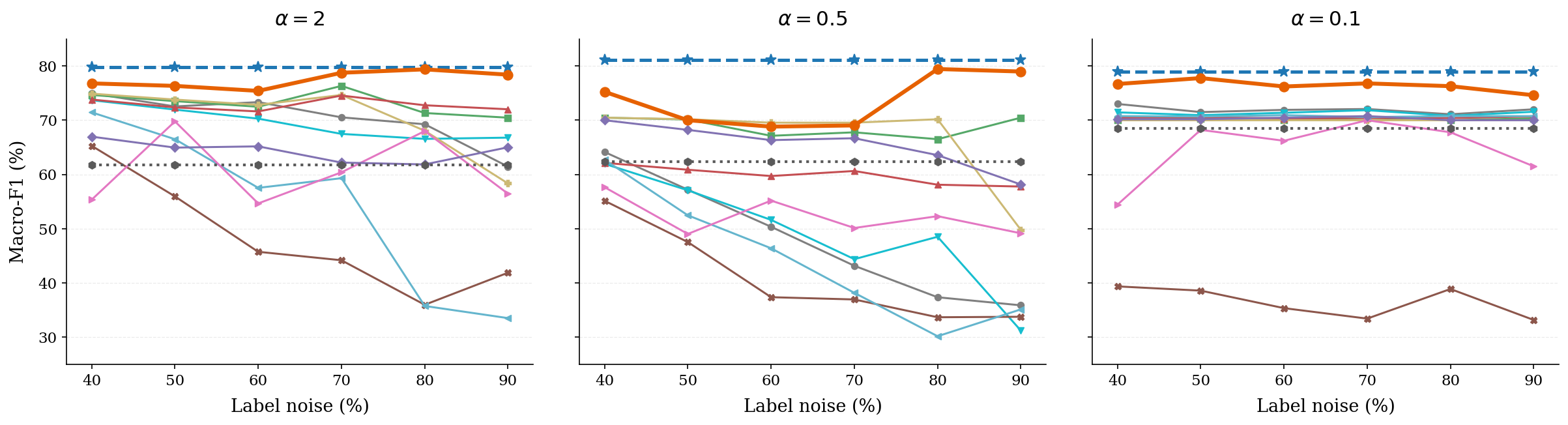}
        \caption{
        WF.
        }
        \label{fig:f1_wf_asym}
    \end{subfigure}


    \caption{
    Macro-F1 performance under asymmetric label noise.
    Results are shown across $\alpha \in \{2.0,0.5,0.1\}$ for
    \textbf{a}, Kermany OCT classification;
    \textbf{b}, UIC DR classification; and
    \textbf{c}, WF GA classification.
    The dashed line denotes the all-clean FedAvg reference.
    }
    \label{fig:macro_f1_asymmetric}

\end{figure}

On Kermany, OCT-FedSIR achieved mean accuracies of 94.76\% and 93.59\% under symmetric and asymmetric noise, respectively. The corresponding values were 88.39\% and 89.57\% for UIC and 75.79\% and 78.47\% for WF. The all-clean FedAvg reference achieved an overall mean accuracy of 91.76\%, a 5.03 percentage-point difference relative to OCT-FedSIR under corrupted supervision.
In addition to overall accuracy, we examined macro-F1 to characterize performance across diagnostic classes. Macro-F1 results under symmetric and asymmetric label noise are shown in
Figs.~\ref{fig:macro_f1_symmetric} and~\ref{fig:macro_f1_asymmetric},
respectively. Across the three ophthalmic tasks, the macro-F1 results generally followed the trends observed for classification accuracy. OCT-FedSIR maintained comparatively strong macro-F1 across increasing label noise levels,
with the greatest degradation occurring when severe symmetric label noise was combined with strong client heterogeneity.

\subsection*{Spectral identification and annotation recovery}
\label{subsec:relabeling_results}
\begin{figure}[!p]
    \centering


    \begin{subfigure}[t]{0.32\textwidth}
        \centering
        \includegraphics[
            width=\linewidth
        ]{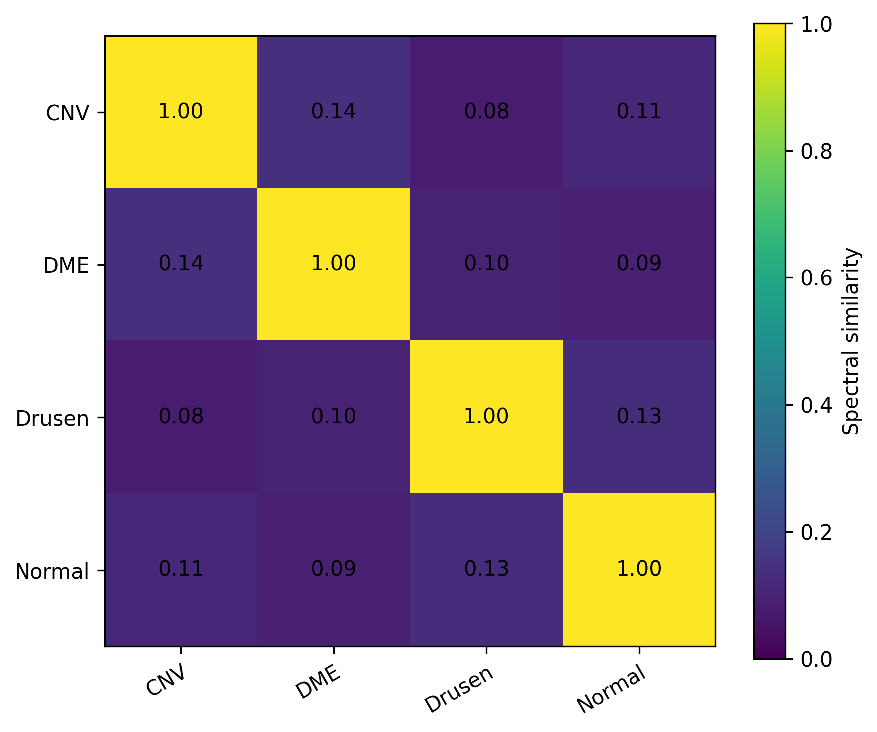}
        \caption{Clean clients}
        \label{fig:client_id_clean_matrix}
    \end{subfigure}
    \hfill
    \begin{subfigure}[t]{0.32\textwidth}
        \centering
        \includegraphics[
            width=\linewidth
        ]{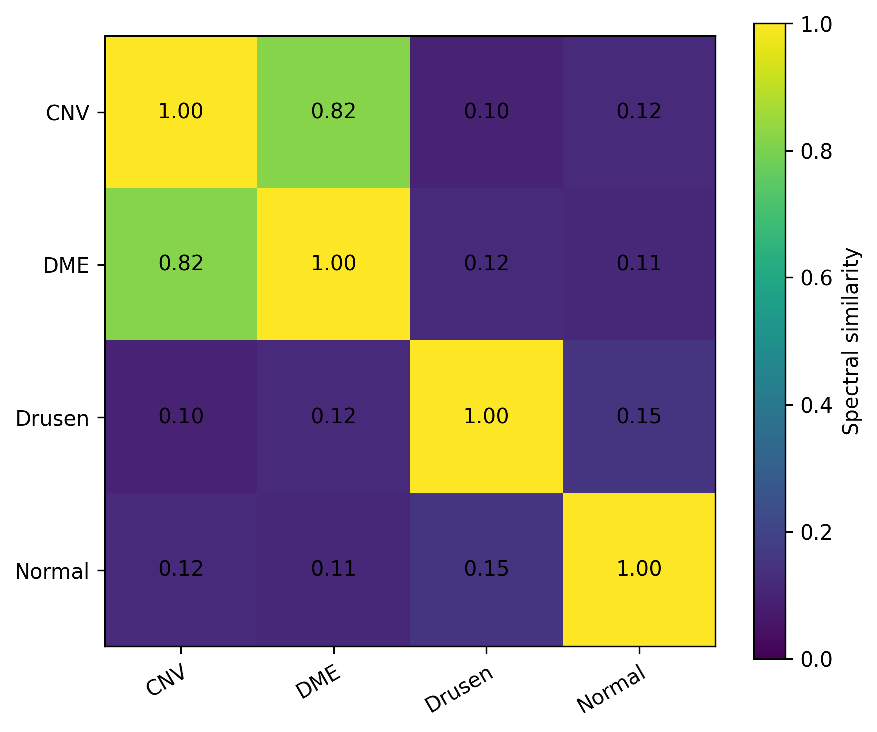}
        \caption{Asymmetric label noise}
        \label{fig:client_id_asym_matrix}
    \end{subfigure}
    \hfill
    \begin{subfigure}[t]{0.32\textwidth}
        \centering
        \includegraphics[
            width=\linewidth
        ]{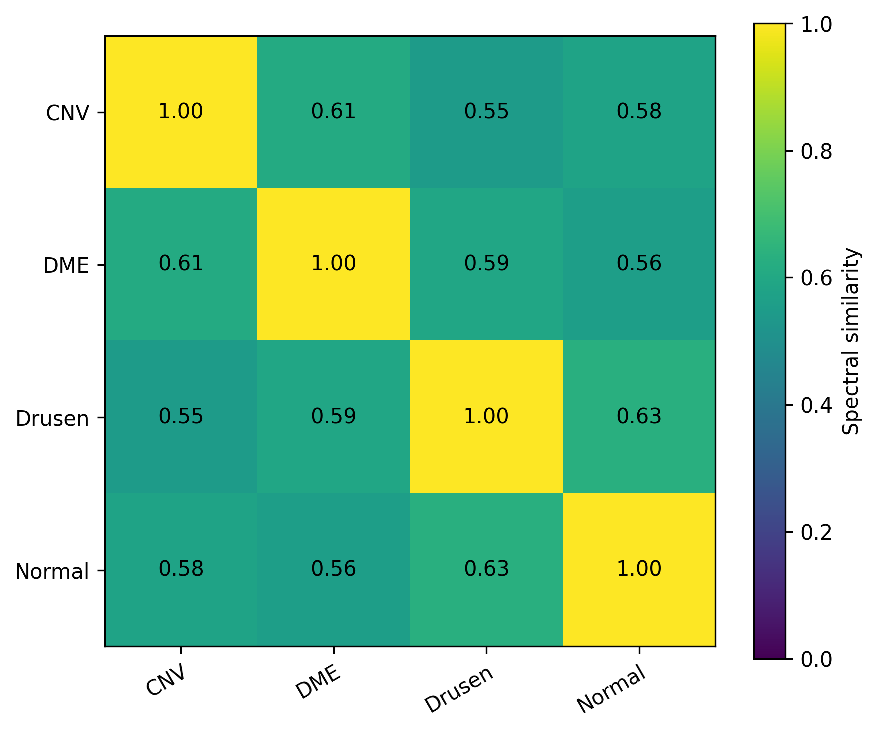}
        \caption{Symmetric label noise}
        \label{fig:client_id_sym_matrix}
    \end{subfigure}

    \vspace{0.7em}


    \begin{subfigure}[t]{0.70\textwidth}
        \centering
        \includegraphics[
            width=\linewidth
        ]{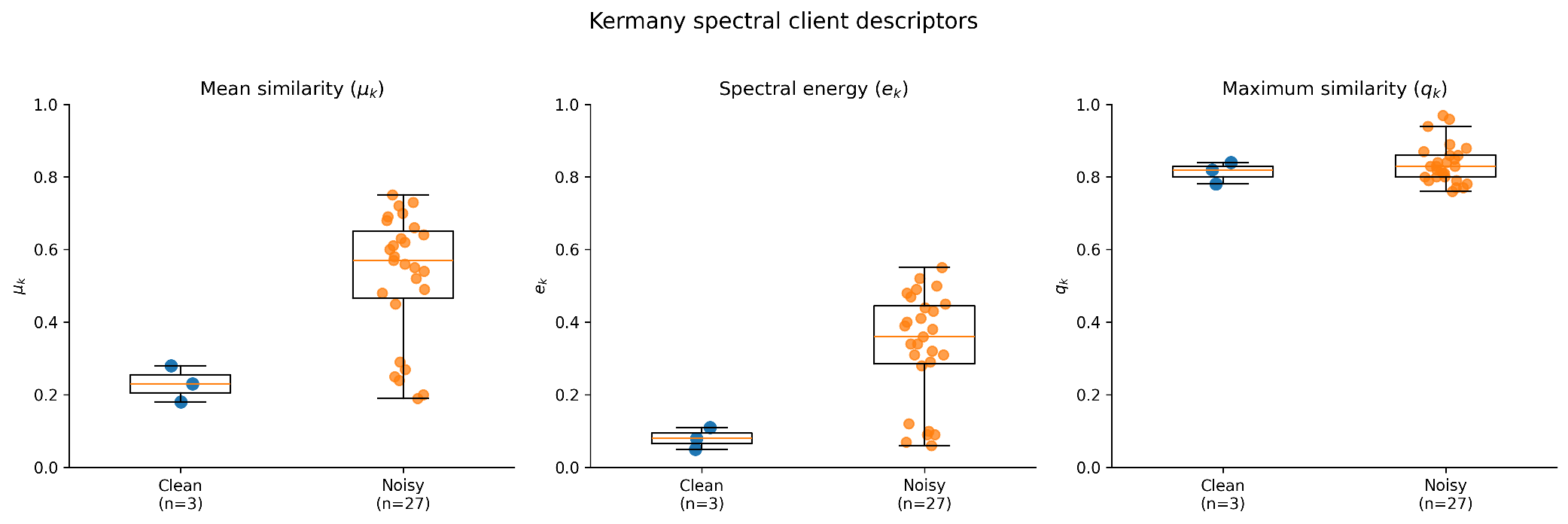}
        \caption{
        Distributions of the client-level spectral descriptors
        $\mu_k$, $e_k$, and $s^{\max}_k$ for clean and noisy clients.
        }
        \label{fig:client_id_distributions}
    \end{subfigure}
    \hfill
    \begin{subfigure}[t]{0.28\textwidth}
        \centering
        \includegraphics[
            width=\linewidth
        ]{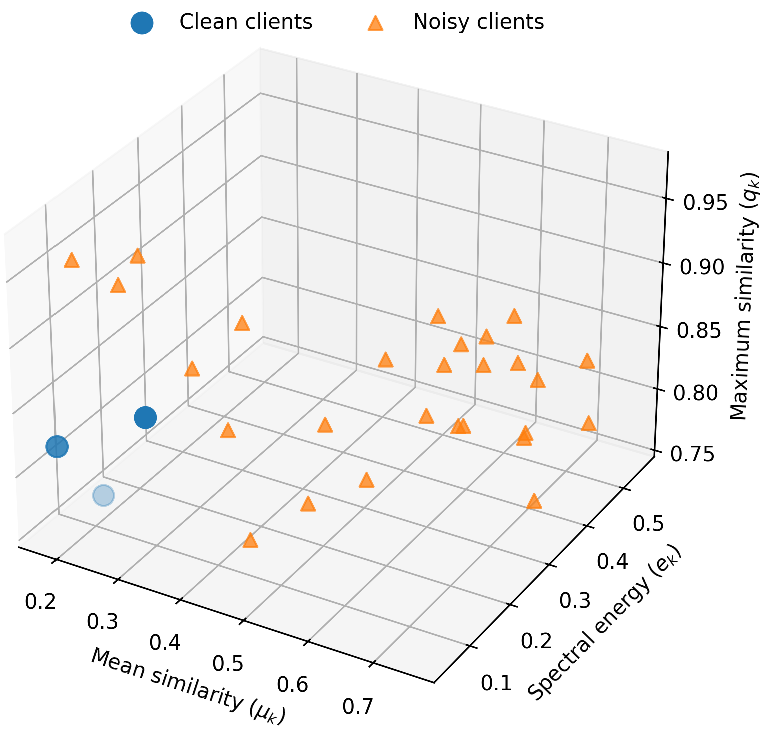}
        \caption{
        Joint spectral descriptor space formed by
        $(\mu_k,e_k,s^{\max}_k)$.
        }
        \label{fig:client_id_joint_space}
    \end{subfigure}

    \vspace{0.7em}


    \begin{subfigure}[t]{1.0\textwidth}
        \centering
        \includegraphics[
            width=\linewidth
        ]{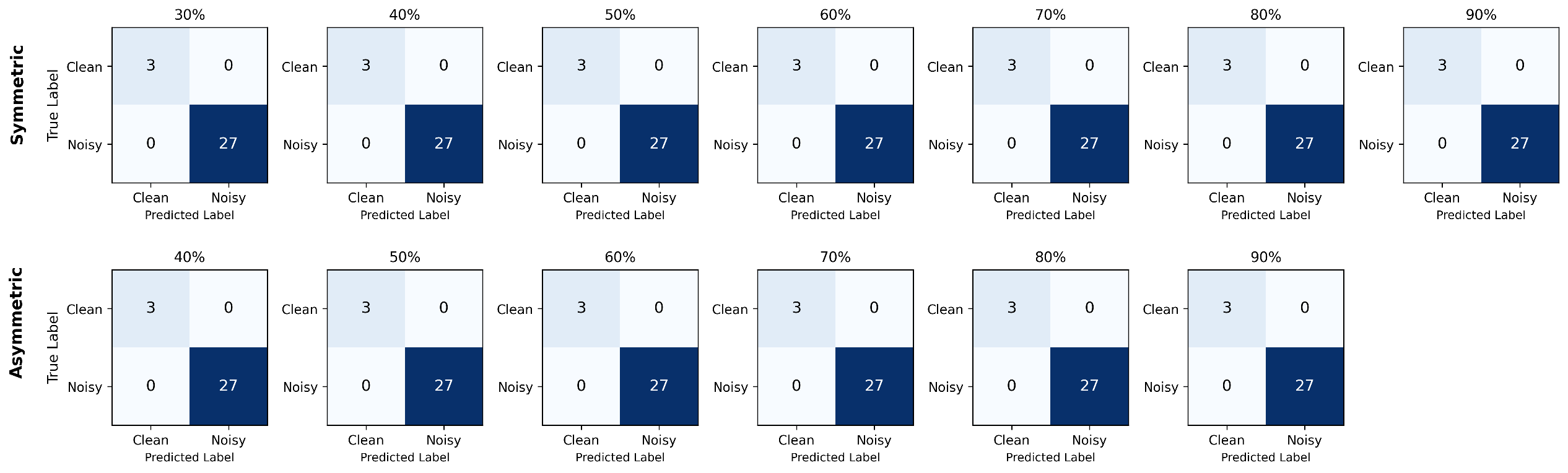}
        \caption{
        Client-identification confusion matrices.
        }
        \label{fig:client_id_pooled}
    \end{subfigure}

    \caption{
    Spectral characterization and identification of clients with corrupted labels
    in representative Kermany experiments under severe statistical heterogeneity
    ($\alpha=0.1$).
    \textbf{a--c}, Class-wise spectral similarity matrices for clean,
    asymmetric-noise, and symmetric-noise clients, respectively.
    \textbf{d}, Distributions of mean off-diagonal similarity ($\mu_k$),
    off-diagonal energy ($e_k$), and maximum off-diagonal similarity
    ($s_k^{\max}$).
    \textbf{e}, Joint spectral descriptor space.
    \textbf{f}, Client-identification confusion matrices across the evaluated
    symmetric and asymmetric noise levels.
    }
    \label{fig:spectral_client_identification}

\end{figure}

We next evaluated the two stages that precede noise-aware federated optimization: identification of clients with corrupted labels and recovery of their sample-level supervision. We first examined whether the FedSIR spectral identification procedure transferred directly to the ophthalmic setting considered here, particularly under structured asymmetric label noise. For this comparison, FedSIR and OCT-FedSIR procedures were evaluated using the same client partitions, corrupted labels, model architectures, and experimental seeds.
The FedSIR identification procedure remained comparatively effective under symmetric noise but showed reduced clean-client F1 under asymmetric noise. The principal failure mode was over-assignment to the clean component: true clean clients were generally retained, but additional clients affected by label noise were also classified as clean. This reduced the purity of the clean reference subset used for subsequent spectral construction. In contrast, the OCT-FedSIR identification procedure completely separated clients retaining their original annotations from clients subjected to simulated label corruption across both noise mechanisms (Table~\ref{tab:original_extended_identification}).
\begin{table}[t]
\centering
\caption{
Clean-client identification F1 score (\%) for FedSIR and OCT-FedSIR,
averaged across the evaluated label-noise levels and Dirichlet heterogeneity
settings. Bold indicates the best-performing method.
}
\label{tab:original_extended_identification}

\begin{tabular}{llcc}
\toprule
Dataset & Method & Symmetric F1 (\%) & Asymmetric F1 (\%) \\
\midrule

Kermany
& FedSIR & 92.4 & 71.8 \\
& OCT-FedSIR & \textbf{100.0} & \textbf{100.0} \\

\addlinespace

UIC
& FedSIR & 89.7 & 68.5 \\
& OCT-FedSIR & \textbf{100.0} & \textbf{100.0} \\

\addlinespace

WF
& FedSIR & 91.2 & 74.6 \\
& OCT-FedSIR & \textbf{100.0} & \textbf{100.0} \\

\bottomrule
\end{tabular}
\end{table}
Across all evaluated datasets, noise mechanisms, noise ratios, and client-heterogeneity settings, OCT-FedSIR correctly recovered the simulated annotation-status partition. Identification accuracy was 100\%, with 100\% recall for clients subjected to label corruption and 100\% precision for clients assigned to the clean reference subset. These results apply to the controlled corruption settings evaluated here.
(Fig.~\ref{fig:spectral_client_identification} shows that complete separation was retained under both symmetric and asymmetric label noise, including the most heterogeneous configuration $\alpha=0.1$).

Compared with the original identification formulation, the extended procedure incorporated class-balanced spectral estimation, Stage-I LA, and the complementary maximum off-diagonal similarity descriptor $s_k^{\max}$. The resulting spectral representations showed greater
separation between clients with clean and corrupted labels. Corrupted-label clients exhibited greater cross-class alignment, reflected by the mean off-diagonal similarity $\mu_k$, off-diagonal energy $e_k$, and maximum off-diagonal similarity $s_k^{\max}$, which jointly formed the descriptor space used by the two-component Gaussian Mixture model (GMM).

We then examined whether identifying corrupted-label clients resulted in recovery of their local annotations. Across the six dataset--noise summaries, spectral relabeling restored a mean of 77.2\% of intentionally corrupted labels to their reference class (Table~\ref{tab:relabeling_summary}). Mean recovery was 75.7\% under symmetric noise and 78.7\% under asymmetric noise. Recovery rates under symmetric and asymmetric noise were 81.6\% and 84.3\% for Kermany,
75.2\% and 78.6\% for UIC, and 70.4\% and 73.1\% for WF, respectively.

Accepted corrections were predominantly consistent with the reference labels. Correction precision ranged from 86.7\% to 95.0\%, with the highest values observed on Kermany. Across the six dataset--noise summaries, the mean
false-correction rate was 3.5\%, corresponding to preservation of approximately 96.5\% of annotations that were initially correct. False-correction rates ranged from 2.4\% to 4.8\%.

The proportion of annotations modified by OCT-FedSIR varied according to dataset and noise mechanism. Under symmetric noise, edit rates were 52.0\%, 49.7\%, and 48.7\% for Kermany, UIC, and WF, respectively. Under asymmetric
noise, the corresponding values were 57.7\%, 55.5\%, and 31.1\%. The lower edit rate observed for WF under asymmetric noise was accompanied by a recovery rate of 73.1\% and correction precision of 89.2\%, demonstrating
that a lower overall frequency of label changes did not preclude recovery of a substantial fraction of corrupted annotations.
\begin{table}[t]
\centering
\caption{
Sample-level spectral relabeling performance at clients selected for
correction. Values are reported as mean $\pm$ standard deviation across three
independent seeds and averaged across the evaluated label-noise levels and
Dirichlet heterogeneity settings. For WF under asymmetric noise, the target
corruption ratio applies only to the GA source class.
}
\label{tab:relabeling_summary}

\begin{adjustbox}{max width=\linewidth}
\begin{tabular}{lccccc}
\toprule
Dataset &
\makecell{Edit rate\\(\%)} &
\makecell{Recovery\\(\%)} &
\makecell{Correction\\precision (\%)} &
\makecell{False-correction\\rate (\%)} &
\makecell{Residual noise\\(\%)} \\
\midrule

\multicolumn{6}{l}{
\textbf{a. Symmetric noise} (mean target noise $=60\%$)
} \\
\addlinespace[2pt]

Kermany &
$52.0 \pm 1.8$ &
$81.6 \pm 2.1$ &
$94.1 \pm 0.9$ &
$2.7 \pm 0.4$ &
$12.1 \pm 1.5$ \\

UIC &
$49.7 \pm 2.2$ &
$75.2 \pm 2.7$ &
$90.8 \pm 1.2$ &
$3.9 \pm 0.6$ &
$16.4 \pm 2.2$ \\

WF &
$48.7 \pm 2.6$ &
$70.4 \pm 3.1$ &
$86.7 \pm 1.5$ &
$4.8 \pm 0.7$ &
$19.7 \pm 2.7$ \\

\addlinespace[4pt]
\midrule

\multicolumn{6}{l}{
\textbf{b. Asymmetric noise} (mean target noise $=65\%$)
} \\
\addlinespace[2pt]

Kermany &
$57.7 \pm 2.0$ &
$84.3 \pm 1.9$ &
$95.0 \pm 0.8$ &
$2.4 \pm 0.4$ &
$11.0 \pm 1.5$ \\

UIC &
$55.5 \pm 2.4$ &
$78.6 \pm 2.4$ &
$92.1 \pm 1.1$ &
$3.4 \pm 0.5$ &
$15.1 \pm 2.0$ \\

WF &
$31.1 \pm 2.8$ &
$73.1 \pm 2.9$ &
$89.2 \pm 1.4$ &
$3.7 \pm 0.6$ &
$12.5 \pm 1.7$ \\

\bottomrule
\end{tabular}
\end{adjustbox}

\end{table}

Together, the client- and sample-level analyses show that the spectral procedure not only identified clients affected by simulated annotation noise but also recovered a substantial fraction of the corrupted supervision while producing comparatively few harmful changes to labels that were originally correct.

\subsection*{Dataset-specific robustness}
\subsubsection*{Kermany}
OCT-FedSIR remained robust across the Kermany experiments, with differences among methods becoming most apparent under severe label noise (Tables~\ref{tab:kermany_sym} and~\ref{tab:kermany_asym}; Figs.~\ref{fig:f1_kermany_sym} and~\ref{fig:f1_kermany_asym}).
The macro-F1 curves showed a pattern consistent with the accuracy results, with OCT-FedSIR maintaining strong class-balanced performance across most symmetric and asymmetric noise settings. Under symmetric noise with $\alpha=2$, OCT-FedSIR achieved 96.00\% accuracy at 80\% noise and 93.60\% at 90\%, compared with 83.50\% for RoFL, 20.60\% for FedAvg, and 19.70\% for FedProx at the 90\% noise level. With $\alpha=0.5$, OCT-FedSIR remained between 97.30\% and 99.70\% across the complete 30--90\% symmetric noise range. Performance decreased under the most heterogeneous configuration. At $\alpha=0.1$, OCT-FedSIR achieved 74.10\% and 74.20\% accuracy at 80\% and 90\% symmetric noise, respectively, compared with 72.60\% and 72.90\% for RoFL. A similar convergence among the methods was observed in macro-F1 at the most severe symmetric-noise levels (Fig.~\ref{fig:f1_kermany_sym}). OCT-FedSIR showed greater stability under asymmetric noise, which was also reflected in macro-F1 (Fig.~\ref{fig:f1_kermany_asym}). With
$\alpha=2$, accuracy remained between 96.70\% and 98.60\% across all evaluated noise levels and reached 97.10\% at 90\% noise, compared with 91.20\% for FedCorr and 71.90\% for RoFL. At $\alpha=0.5$, OCT-FedSIR retained 95.70\% accuracy at 90\% noise. Under the strongest heterogeneity ($\alpha=0.1$), accuracy decreased from 95.50\% at 40\% noise to 75.60\% at 90\%, but remained above RoFL (71.50\%), FedNoRo (64.10\%), and FedLSR (60.10\%) at the highest noise level.

\subsubsection*{UIC}
OCT-FedSIR achieved the highest accuracy among the evaluated noisy-label methods across the reported UIC symmetric and asymmetric conditions (Tables~\ref{tab:uic_sym} and~\ref{tab:uic_asym}). The corresponding macro-F1 results showed similar robustness across increasing noise levels
(Figs.~\ref{fig:f1_uic_sym} and~\ref{fig:f1_uic_asym}). Under symmetric noise with $\alpha=2$, accuracy decreased from 95.80\% at 30\% noise to 84.60\% at 90\%, compared with 72.30\% for RoFL and 68.80\% for FedCorr at
the highest noise level. With $\alpha=0.5$, OCT-FedSIR retained 80.20\% accuracy at 90\% noise, compared with 71.60\% for FedCorr and 68.20\% for RoFL. At $\alpha=0.1$, OCT-FedSIR decreased from 89.60\% at 30\% symmetric noise to 73.80\% at 90\%, whereas RoFL achieved 69.10\% at the highest noise level. Under asymmetric noise, OCT-FedSIR remained comparatively stable in both accuracy and macro-F1 (Fig.~\ref{fig:f1_uic_asym}). With $\alpha=2$, OCT-FedSIR remained above 90\% accuracy throughout the 40--90\% noise range and achieved 90.70\% at 90\% noise, compared with 80.50\% for FedCorr and 72.40\% for RoFL. At the same 90\% noise level, OCT-FedSIR achieved 88.60\% with $\alpha=0.5$ and 81.40\% with $\alpha=0.1$.

\subsubsection*{WF}
The differences among methods were smaller on WF than on Kermany and UIC in several symmetric-noise settings (Table~\ref{tab:wf_sym}). This behavior was also reflected in macro-F1, where the separation among the stronger methods was narrower than that observed for Kermany and UIC. With $\alpha=2$, OCT-FedSIR remained between 77.57\% and 78.74\% across the complete 30--90\% symmetric noise range. At $\alpha=0.5$, RoFL exceeded OCT-FedSIR at several intermediate noise levels, whereas OCT-FedSIR performed more strongly as noise became severe. At 90\% noise, OCT-FedSIR achieved 72.82\%, compared with 62.44\% for RoFL, 42.30\% for FedAvg, and 33.37\% for FedProx. Under $\alpha=0.1$, OCT-FedSIR achieved 78.91\%, 77.96\%, 77.90\%, and 76.67\% at 30\%, 40\%, 50\%, and 60\% symmetric noise, respectively, and retained 72.99\% at 90\%. RoFL slightly exceeded OCT-FedSIR at 80\% noise (72.99\% versus 72.15\%), whereas OCT-FedSIR achieved the higher accuracy at the remaining symmetric noise levels. A clearer separation was observed under the one-directional GA-to-NGA asymmetric noise (Table~\ref{tab:wf_asym}; Fig.~\ref{fig:f1_wf_asym}). The macro-F1 results similarly showed stronger separation between OCT-FedSIR
and the competing approaches under severe asymmetric noise. At $\alpha=2.0$, OCT-FedSIR achieved the highest accuracy at all six evaluated noise levels, including 81.92\% and 80.97\% at 80\% and 90\% noise, respectively. At $\alpha=0.5$, several competing methods were comparable at intermediate noise levels, but OCT-FedSIR reached 81.98\% and 81.53\% at 80\% and 90\% noise. With $\alpha=0.1$, OCT-FedSIR ranged from 77.29\% to 80.36\% across the evaluated asymmetric noise levels.

\subsection*{Retention versus pruning of clients with corrupted annotations}
The pruning experiment evaluated whether clients identified as having corrupted annotations should be excluded from subsequent federated optimization or retained after spectral correction. Across all 117 evaluated conditions, OCT-FedSIR achieved an overall mean accuracy of 86.73\%, compared with 77.43\% for pruning.
On Kermany, OCT-FedSIR achieved mean accuracies of 94.76\% and 93.59\% under symmetric and asymmetric noise, respectively, compared with 81.23\% and
84.20\% for pruning. These results correspond to improvements of 13.53 percentage points and 9.39 percentage points, respectively.
On UIC, OCT-FedSIR achieved 88.39\% under symmetric noise and 89.57\% under asymmetric noise, compared with 83.47\% and 79.97\% for pruning. The largest difference on UIC was observed under asymmetric noise, where OCT-FedSIR exceeded pruning by 9.60 percentage points. On WF, OCT-FedSIR achieved mean accuracies of 75.79\% and 78.47\% under symmetric and asymmetric noise, respectively, whereas pruning achieved 67.22\% and
68.53\%. The corresponding differences were 8.57 and 9.94 percentage points.

\subsection*{Ablation study}
We used a one-component-at-a-time removal analysis to assess the sensitivity of the complete OCT-FedSIR framework to relabeling, KD, LA, and DaAgg (Table~\ref{tab:OCT-FedSIR_ablation}). Removing relabeling produced the largest decrease in performance, reducing mean accuracy from 86.73\% to 79.47\%, a difference of 7.26 percentage points. Removing KD, LA, and DaAgg reduced mean accuracy by 1.82, 1.05, and 0.52 percentage points, respectively. Thus, performance was most sensitive to removal of the relabeling component, while KD, LA, and DaAgg provided additional gains within the complete framework.
\begin{table}[t]
\centering
\caption{
Ablation analysis of OCT-FedSIR. Mean accuracy is averaged across all
evaluated experimental conditions. $\Delta$ denotes the decrease from the
complete framework in percentage points.
}
\label{tab:OCT-FedSIR_ablation}

\begin{tabular}{lcc}
\toprule
Configuration & Mean accuracy (\%) & $\Delta$ \\
\midrule

\textbf{OCT-FedSIR (Full)} & \textbf{86.73} & -- \\
w/o DaAgg              & 86.21 & $-0.52$ \\
w/o LA                 & 85.68 & $-1.05$ \\
w/o KD                 & 84.91 & $-1.82$ \\
w/o relabeling         & 79.47 & $-7.26$ \\

\bottomrule
\end{tabular}
\end{table}

\section*{Discussion}
This study addressed an underexplored challenge for trustworthy ophthalmic FL: whether annotation-related disruption can be distinguished from legitimate differences in the clinical data distributions of participating clients. Across 117 dataset-specific experimental conditions spanning annotation-noise mechanism, noise severity, and client heterogeneity, OCT-FedSIR achieved an overall mean accuracy of 86.73\%. This was 6.79 percentage points higher than RoFL and 7.98 percentage points higher than FedCorr, the two strongest comparison methods overall, and 5.03 percentage points below the all-clean FedAvg reference. Macro-F1 showed broadly similar trends across increasing noise levels. These findings indicate that explicitly modeling annotation reliability can substantially reduce the degradation associated with corrupted supervision while preserving heterogeneous client data within federated optimization.

An important finding was that the original FedSIR spectral identification strategy did not transfer uniformly to the ophthalmic setting. Although FedSIR remained comparatively effective under symmetric label-noise, its clean-client F1 decreased substantially under structured asymmetric noise, ranging from 68.5\% to 74.6\% across the three datasets. The principal failure mode was the assignment of some label-corrupted clients to the clean reference group. In contrast, OCT-FedSIR correctly separated the simulated clean and label-corrupted clients across all evaluated datasets, noise mechanisms, noise levels, and heterogeneity settings. The combined identification formulation incorporates class-balanced spectral estimation, Stage-I LA, and the complementary maximum off-diagonal similarity descriptor $s_k^{\max}$. These results support the importance of adapting spectral client characterization to the class-distribution heterogeneity encountered in ophthalmic data rather than assuming that a spectral identification strategy developed on general vision benchmarks will transfer unchanged.

The complete client-identification result should nevertheless be interpreted within the controlled experimental setting. Client status was defined through simulated corruption of existing reference annotations, and the resulting clean and noisy populations may exhibit more structured separation than would
occur across independent clinical institutions. The finding therefore demonstrates that OCT-FedSIR distinguished the corruption patterns evaluated here; it should not be interpreted as evidence that annotation reliability can be identified perfectly in prospective multicenter data. Naturally occurring annotation disagreement may coexist with scanner shifts, disease-prevalence differences, image-quality variation, referral patterns, and institutional practice differences, all of which may alter feature geometry independently of annotation quality.

Client identification was also only the first part of the problem. At the sample level, spectral relabeling recovered a mean of 77.2\% of intentionally corrupted annotations across the six dataset--noise summaries. Correction precision ranged from 86.7\% to 95.0\%, while the mean false-correction rate was 3.5\%, corresponding to preservation of approximately 96.5\% of annotations that were originally correct. This distinction is important because improved classification performance alone does not establish that an annotation-correction strategy is behaving safely. The observed combination of substantial recovery and comparatively infrequent harmful changes is consistent with the agreement rule used by OCT-FedSIR, in which a label is modified only when the representative and residual-subspace criteria support the same alternative class.

The interaction between annotation noise and statistical heterogeneity remained apparent despite the strong client-identification performance. Under the most concentrated client distributions, particularly at $\alpha=0.1$, classification performance decreased in several severe-noise conditions. This indicates that correctly identifying a label-corrupted client does not guarantee that its annotations can subsequently be corrected with equal reliability. When local class support becomes sparse or highly concentrated, the class-specific structure required to construct and apply reliable spectral references may itself become less stable. This distinction is clinically relevant because a specialized referral center may contain an unusual or highly skewed disease distribution despite having reliable annotations. Such legitimate variation should not itself be interpreted as evidence of poor data quality.

The structured asymmetric experiments provide an additional perspective on this problem. Corruption was concentrated between selected diagnostic categories rather than distributed uniformly: between choroidal neovascularization (CNV) and diabetic macular edema (DME) and between drusen and normal retina for Kermany, between neighboring DR severity categories for UIC, and from GA to NGA for WF. These transitions were designed as controlled structured perturbations and should not be interpreted as estimates of real clinical error probabilities. OCT-FedSIR remained particularly robust under these asymmetric settings, including severe label-noise on Kermany and UIC and the one-directional WF experiment. This is notable because annotation disagreement in clinical practice is unlikely to be uniform across diagnostic categories and may instead concentrate near phenotypically or clinically related decision boundaries.

Differences across the three tasks further suggest that spectral robustness is dependent on the structure of the classification problem. Kermany and UIC provide multiple class relationships from which cross-class spectral structure can be characterized, whereas the binary WF task provides fewer such relationships. Correspondingly, differences among the stronger methods were smaller in several WF symmetric-noise settings, although OCT-FedSIR retained a clearer advantage under severe asymmetric noise. These findings caution against interpreting performance differences among the three datasets as differences in dataset difficulty alone, particularly because the tasks, cohorts, and network architectures also differed.

The pruning comparison addressed a complementary question: whether a client identified as having corrupted supervision should be excluded or retained after correction. Across all evaluated conditions, OCT-FedSIR exceeded spectral pruning by 9.30 percentage points on average. This result highlights an important distinction between annotation reliability and data value. A client with corrupted annotations may still contain correctly labeled samples, uncommon disease manifestations, or other clinically informative variation that would be lost if the entire client were excluded. Retaining and correcting such clients was therefore generally more effective than discarding them, although this advantage was not universal. Under the most heterogeneous Kermany setting with severe asymmetric corruption, pruning outperformed correction, suggesting a boundary at which the available spectral evidence may no longer support dependable recovery.

The component ablation provides additional support for this interpretation. Removing spectral relabeling from the complete framework produced the largest decrease in mean accuracy, from 86.73\% to 79.47\%, a reduction of 7.26 percentage points. Removing KD, LA, and DaAgg produced smaller decreases of 1.82, 1.05, and 0.52 percentage points, respectively. These comparisons do not isolate independent causal effects because the components operate jointly; rather, they show that relabeling was the component whose removal from the complete framework was associated with the largest performance loss. KD provided complementary supervision when hard labels remained uncertain, LA addressed local class imbalance, and DaAgg reduced the influence of label-corrupted client models that deviated from the clean reference population. OCT-FedSIR therefore acts at multiple levels, combining client identification, sample-level supervision correction, local optimization, and server aggregation.

This study has several limitations. OCT-FedSIR depends on sufficient reliable class support for constructing its spectral references, and performance may deteriorate when too few reliable clients remain or when diagnostic classes are absent or sparsely represented. Without a sufficiently reliable spectral reference, however, the framework cannot be expected to distinguish annotation-related deviations consistently or to support dependable relabeling of corrupted supervision. In addition, robustness to annotation noise does not by itself establish clinical utility or formal privacy. Client-identification outputs should be interpreted as signals of potential data-quality concerns, and spectrally corrected labels as training targets rather than changes to clinical records. Independent-site validation is therefore needed before clinical use, while future work should assess information-leakage risks from model updates and shared spectral statistics, including compatibility with secure aggregation and differential privacy.

In summary, OCT-FedSIR addressed annotation reliability as a distinct challenge in ophthalmic FL by coupling spectral client characterization with selective recovery of corrupted supervision. Under the controlled conditions evaluated here, OCT-FedSIR substantially improved separation of clients with reliable and corrupted annotations relative to the original FedSIR procedure, while spectral relabeling recovered a large proportion of corrupted annotations with relatively few harmful changes to labels that were already correct. Retaining corrected clients was generally more effective than discarding them, further demonstrating that annotation reliability and data value should not be treated as equivalent. Together, these findings identify annotation reliability as an important requirement for trustworthy ophthalmic FL and motivate validation under naturally occurring annotation disagreement and genuinely multicenter clinical data.


\section*{Methods}
\label{sec:methodology}

\subsection*{Study design}

We evaluated FL under two forms of heterogeneity that can coexist across clinical sites: differences in local disease composition and differences in annotation reliability. For each experimental seed, training data were partitioned among simulated clients using a class-aware Dirichlet procedure. A subset of clients retained the original labels, whereas labels at the remaining clients were corrupted using controlled symmetric or asymmetric transition rules. All methods within an experimental condition used the same client partition, noisy-client assignment, noisy labels, model initialization, and training budget. Three dataset-specific lightweight backbones were used, one for each retinal classification task.

OCT-FedSIR comprised three stages: spectral identification of clients with reliable and unreliable labels (Fig.~\ref{fig:stages}A--E), periodic relabeling of identified noisy clients using spectral references constructed from clean clients (Fig.~\ref{fig:stages}F), and noise-aware federated optimization using LA, KD, and DaAgg (Fig.~\ref{fig:stages}G).

\subsection*{Datasets}

Experiments used the Kermany OCT dataset, a UIC DR cohort, and a WF GA cohort.

\subsubsection*{Kermany}

The Kermany OCT2017 dataset is a publicly available spectral-domain OCT dataset introduced by Kermany et al. \cite{kermany}. It contains B-scans acquired with Heidelberg Spectralis OCT systems and labeled into four diagnostic categories: CNV, DME, drusen, and normal retina. Duplicate images were removed before federated partitioning. The original held-out test set was kept separate from all federated training procedures.

\subsubsection*{UIC}
The UIC cohort contains OCT imaging from patients categorized as control, mild, moderate, or severe DR \cite{nabil2025federated}. The study received institutional review board approval from the University of Illinois Chicago and was conducted in accordance with the Declaration of Helsinki. DR severity was assigned by a retina specialist according to the Early Treatment Diabetic Retinopathy Study grading framework. OCT/OCTA data were acquired with an ANGIOVUE spectral-domain OCTA system (Optovue, Fremont, CA) operating at a 70-kHz A-scan rate, with an axial resolution of approximately $5\,\mu\mathrm{m}$ and a lateral resolution of approximately $15\,\mu\mathrm{m}$. The cohort contained 445 scans from 41 patients that satisfied a signal-strength threshold of $Q \geq 5$, including 187 3-mm scans and 258 6-mm scans. For the present classification experiments, patient-level separation was maintained between development and held-out evaluation sets so that images from the same patient did not occur in more than one split. Individual OCT B-scans were used as model inputs for the classification task.

\subsubsection*{Wake Forest}
The WF cohort was collected at Wake Forest University School of Medicine between January 1, 2013, and January 28, 2023. The study protocol received institutional review board approval and adhered to the Declaration of Helsinki. Patients were identified using ICD-9 codes 362.50 and 362.51 and ICD-10 codes H35.30 and H35.31 and were included when longitudinal OCT imaging was available with sufficient follow-up to assess GA progression. Exclusion criteria included subfoveal GA at baseline, fewer than two available OCT visits, unavailable longitudinal imaging, prior retinal treatment, and ocular conditions likely to confound OCT interpretation, including neovascular AMD, DR, glaucoma, retinal vascular occlusion, retinal detachment, and macular dystrophy. After application of these criteria, 91 patients with adequate imaging were retained. A single predefined study eye was used for each patient and remained fixed across visits (OD, 48.08\%; OS, 51.92\%). The cohort contained 455 OCT volumes and 11,513 B-scans: 258 volumes with central GA (CGA; 6,522 B-scans), 74 with non-central GA (NCGA; 1,874 B-scans), and 123 with no GA (NGA; 3,117 B-scans). CGA and NCGA were combined into a single GA category for the present study. This formulation separates the presence from the absence of GA without introducing the additional localization-dependent distinction between central and non-central atrophy. Individual OCT B-scans were used as model inputs for the binary classification task. All partitions were performed at the patient level before extraction of individual B-scans; no B-scan, OCT volume, eye, or visit from a patient in the held-out test set appeared in the federated training data.

The three datasets therefore represented different classification settings: multiclass OCT disease classification in Kermany, ordinal DR severity classification in UIC, and binary GA classification in WF. Each sample retained a unique global index so that the original label, simulated corrupted label, and any subsequent spectral label could be tracked. Ground-truth labels were used only to generate controlled corruption and to evaluate identification and correction; they were not provided to the FL algorithms.

\subsection*{Federated client construction}
Let the federated training set at client k be denoted by
$\mathcal{D}_k = \{(x_i,y_i)\}_{i=1}^{N_k},$ where $N_k$ is the total number of training samples and $y_i\in \{1,\ldots,C\}$ denotes one of $C$ diagnostic classes. The Kermany, UIC, and WF training sets were partitioned among $K=30$, $K=20$, and $K=5$ federated clients, respectively. Only training samples were distributed among clients; the held-out test sets remained completely separate from the federated partitioning procedure. To simulate statistical heterogeneity across clients, the training data were partitioned using a class-wise Dirichlet distribution. For each diagnostic class $c$, the proportions of samples assigned across the $K$ clients were drawn from $
{\rho}_c
\sim
\operatorname{Dirichlet}
\left(
\alpha \mathbf{1}_K
\right),
$ where ${\rho}_c$ represents the allocation proportions of class $c$ across the participating clients and its entries sum to one. Samples belonging to class $c$ were then distributed among clients according to these proportions.

The concentration parameter $\alpha$ controlled the degree of statistical class heterogeneity across clients. We evaluated
$
\alpha
\in
\{2.0,0.5,0.1\},
$
where smaller values of $\alpha$ produce more heterogeneous non-IID client distributions with increasingly uneven local class proportions. For each experimental seed and value of $\alpha$, the client partition was generated once and then held fixed across all evaluated FL methods. Thus, competing methods within the same experimental condition were trained using identical client data partitions.

\subsection*{Client-dependent label-noise simulation}

After the federated client partitions were generated, clients were randomly assigned as either clean or noisy to simulate client-dependent annotation unreliability. 
Across the evaluated datasets and client configurations, the proportion of label-corrupted clients ranged from $60\%$ to $90\%$, with the remaining $10\%$--$40\%$ clean-label clients. The clean/label-corrupted status of each client was unknown a priori.  
The clean/noisy assignment was generated independently for each experimental seed and was independent of the subsequent spectral client-identification procedure. Let $
    \gamma_k \in \{0,1\}
$
denote the simulated annotation status of client $k$, where
\begin{equation}
    \gamma_k =
    \begin{cases}
        0, & \text{if client } k \text{ is clean}, \\[4pt]
        1, & \text{if client } k \text{ is noisy}.
    \end{cases}
\end{equation}

For a given experimental seed, the clean/noisy client assignment was generated once and then held fixed across all competing FL methods. Consequently, methods evaluated within the same experimental condition used the same client partition, clean/noisy client assignment, noisy labels, and model initialization. Ground-truth client status $\gamma_k$ was used only to evaluate client-identification performance; it was not provided to OCT-FedSIR during training or spectral client identification.

Clean clients retained their original annotations. For each client designated as noisy, label-noise was applied separately within every locally represented class. This class-conditional procedure ensured that differences in local class prevalence did not determine the imposed noise level. Let $n_{k,c}$ denote the number of samples belonging to class $c$ at client $k$ before corruption. For a target label-noise ratio $\eta$, the number of samples selected for corruption within class $c$ was $
    \lfloor
        \eta n_{k,c} \rfloor.
$ The corresponding samples were selected uniformly at random without replacement. Thus, $\eta$ controlled the fraction of annotations corrupted within each locally represented class of a noisy client.

\begin{figure}[htbp]
    \centering
    \includegraphics[
        width=\textwidth
    ]{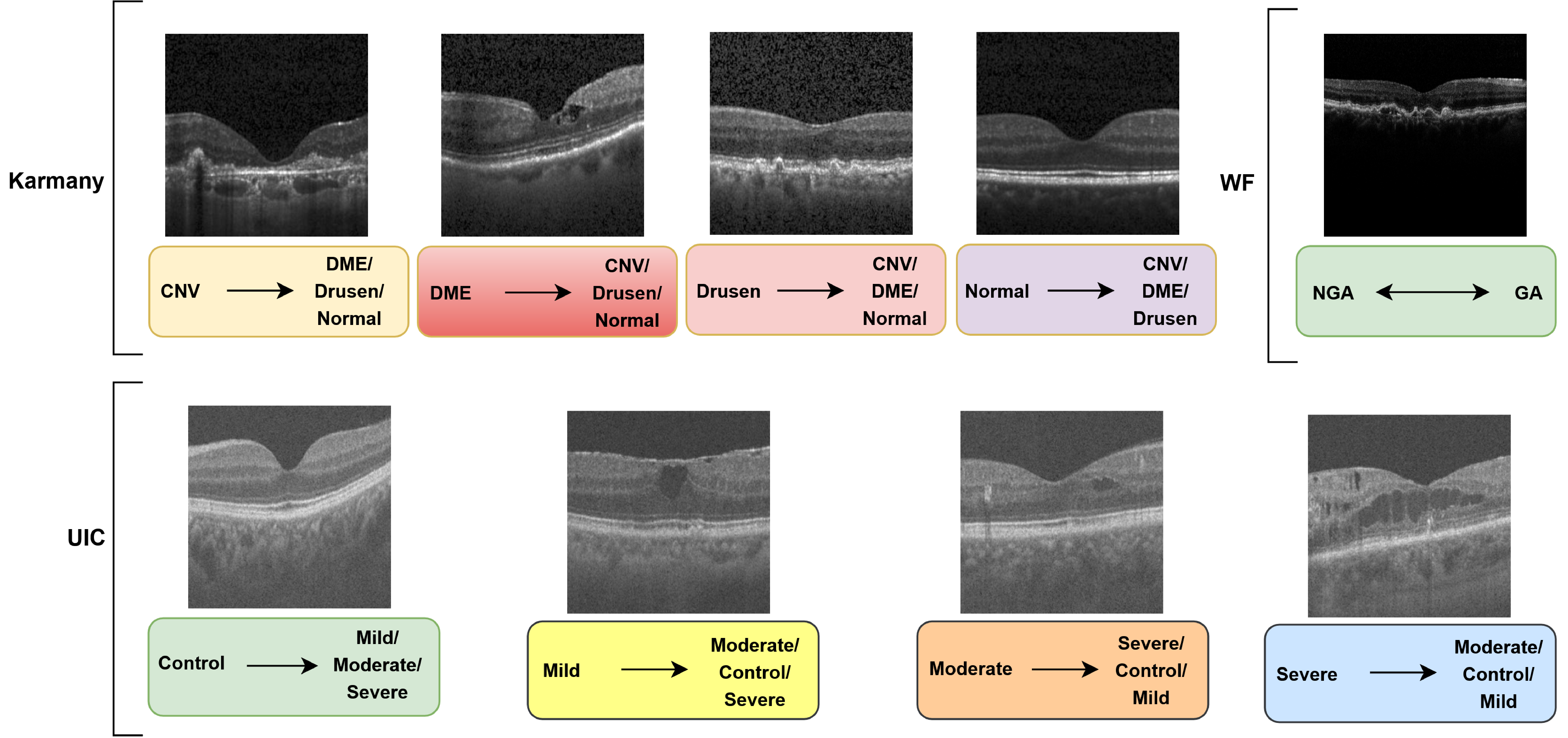}
    \caption{
    Simulated symmetric label-noise transitions. For clients designated as noisy,
    a proportion $\eta$ of annotations within each locally represented class was
    selected uniformly at random and reassigned uniformly to one of the remaining
    $C-1$ diagnostic classes. Clean clients retained their original annotations.
    }
    \label{fig:data_samples_symmetric}
\end{figure}

\begin{figure}[htbp]
    \centering
    \includegraphics[
        width=\textwidth
    ]{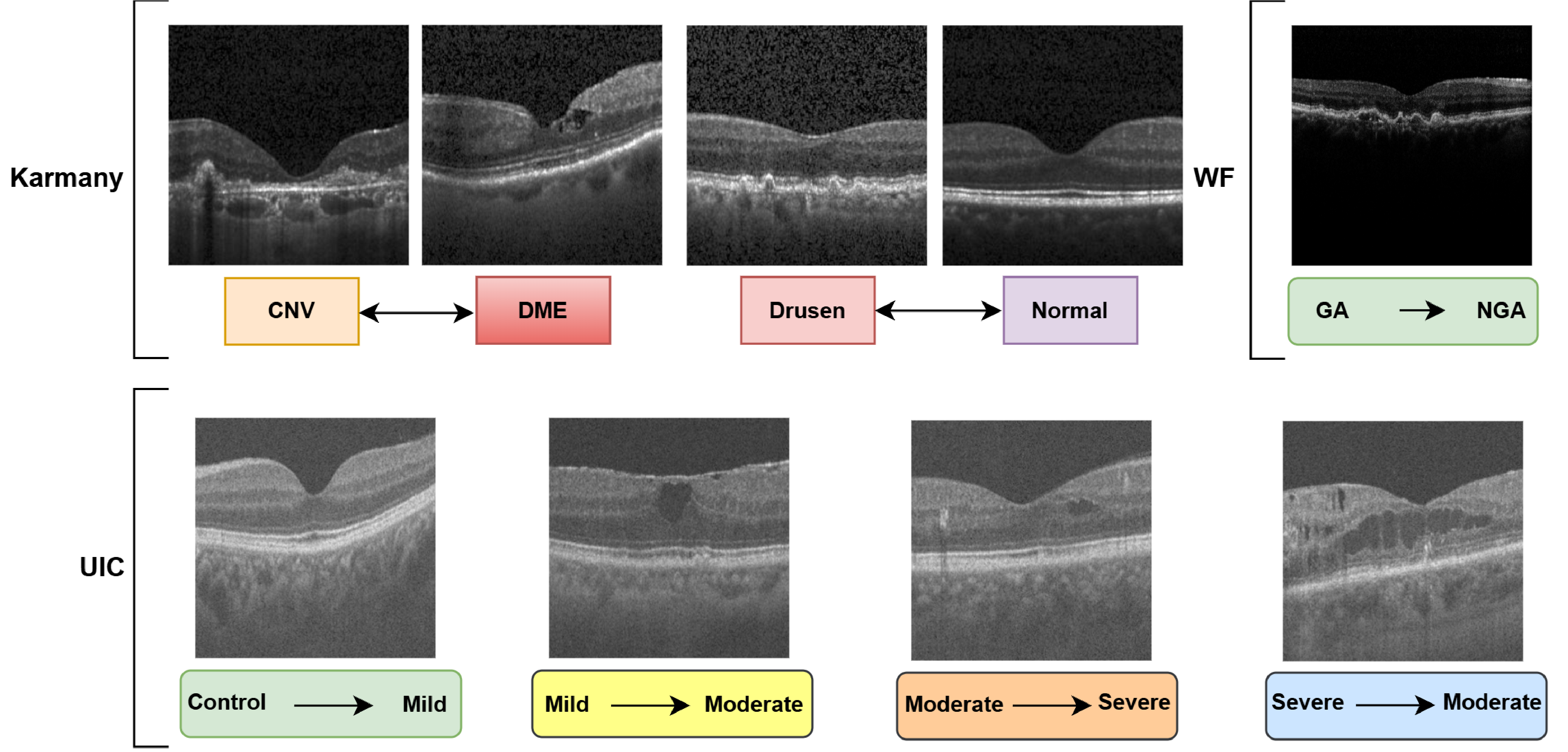}
    \caption{
    Dataset-specific asymmetric label-noise transitions. For clients designated
    as noisy, a proportion $\eta$ of annotations within each applicable source
    class was selected and reassigned according to predefined dataset-specific
    class transitions. Clean clients retained their original annotations.
    }
    \label{fig:data_samples_asymmetric}
\end{figure}

For symmetric label-noise, each selected annotation from true class $c$ was
reassigned uniformly to one of the remaining $C-1$ diagnostic classes
(Fig.~\ref{fig:data_samples_symmetric}). The resulting transition model was

\begin{equation}
    P(\widetilde{y}=j \mid y=c)
    =
    \begin{cases}
        1-\eta, & j=c, \\[4pt]
        \dfrac{\eta}{C-1}, & j\neq c.
    \end{cases}
\end{equation} where $y$ denotes the original reference label and $\widetilde{y}$ denotes the observed label after simulated corruption. Symmetric label-noise ratios were evaluated at $
    \eta_{\mathrm{sym}}
    \in
    \{0.3,0.4,0.5,0.6,0.7,0.8,0.9\}.
$ For asymmetric label-noise, the same within-class sampling procedure was used,
but selected annotations were reassigned according to predefined
dataset-specific class transitions rather than uniformly across all alternative
classes (Fig.~\ref{fig:data_samples_asymmetric}). Let $g(c)$ denote the predefined target class associated with source class $c$. The transition model was therefore

\begin{equation}
    P(\widetilde{y}=j \mid y=c)
    =
    \begin{cases}
        1-\eta, & j=c, \\[4pt]
        \eta, & j=g(c), \\[4pt]
        0, & \text{otherwise}.
    \end{cases}
\end{equation} For the Kermany dataset, asymmetric transitions were introduced between clinically related diagnostic categories, including CNV and DME, and between drusen and normal retina. For UIC, asymmetric transitions were restricted to neighboring DR severity categories to preserve the ordinal structure of the task. For WF, asymmetric corruption followed a predefined one-directional transition from GA to NGA. Asymmetric label-noise ratios were evaluated at $
    \eta_{\mathrm{asym}}
    \in
    \{0.4,0.5,0.6,0.7,0.8,0.9\}.$ The symmetric noise provided a class-independent stress test, whereas asymmetric noise represented structured annotation errors concentrated between selected diagnostic categories.

\subsection*{Class balancing for spectral estimation}

Because non-IID partitioning and label-noise could produce substantial
differences in the number of samples associated with individual classes within a client, a class-balanced loader was constructed for the spectral
client-identification stage of OCT-FedSIR and for the corresponding spectral
pruning baseline. Balancing was performed using the observed, potentially
corrupted labels available to each client and did not use the underlying
ground-truth clean labels. For this balancing procedure, let $n_{k,c}$ denote the number of samples at
client $k$ carrying the observed label $c$, and let
\begin{equation}
    \mathcal{C}_k
    =
    \{c : n_{k,c}>0\}
\end{equation}
denote the set of classes represented by the observed local annotations. The
largest observed class count at client $k$ was defined as
\begin{equation}
    n_k^{\max}
    =
    \max_{c\in\mathcal{C}_k}
    n_{k,c}.
\end{equation}
Each observed class was oversampled with replacement until its sample count
matched $n_k^{\max}$. Thus, for each class $c\in\mathcal{C}_k$ with
$n_{k,c}<n_k^{\max}$,
\begin{equation}
    n_k^{\max}-n_{k,c}
\end{equation}
additional instances were sampled from the existing local examples of that
class. Only classes already represented at a client were included in the balancing
procedure; classes absent from the observed local annotations were not
introduced. Consequently, balancing equalized the number of samples across
the classes present at each client without altering its observed class support.
No additional augmentation specific to the balancing procedure was applied;
class balancing itself was performed solely by randomly repeated sampling of existing
local examples.

The balanced loader was used only for feature extraction and spectral estimation after Stage-I local training; it was not used to optimize the Stage-I client models. Stage-I local optimization was performed using the original client datasets with the LA objective described below. After local training, we used the balanced loader to extract class-specific feature representations from which the spectral descriptors were estimated. All subsequent federated local optimization likewise used the original client datasets.

\subsection*{FL formulation}
At communication round $t$, the server distributed the global parameters $\boldsymbol{\theta}^{(t)}$ to all clients. Each client performed local optimization and returned updated parameters $\boldsymbol{\theta}^{(t)}_k$. Under conventional FedAvg \cite{macmahan}, the global model is updated as
\begin{equation}
    \boldsymbol{\theta}^{(t+1)}
    =
    \sum_{k=1}^{K}
    \frac{N_k}{\sum_{j=1}^{K}N_j}
    \boldsymbol{\theta}^{(t)}_k.
\end{equation}
This aggregation does not explicitly account for differences in annotation reliability. OCT-FedSIR instead estimates a clean-client subset, uses those clients to generate class-specific spectral references, and adjusts both local supervision and server aggregation according to the estimated client status.

\subsection*{OCT-FedSIR for noisy-label ophthalmic FL}

\subsubsection*{Stage-I: spectral client identification}

Because the annotation-quality status of each client is unknown a priori, Stage-I aims to distinguish clients with reliable annotations from clients with potentially corrupted labels. OCT-FedSIR performs this identification using spectral descriptors derived from locally learned class-wise feature representations, without access to ground-truth client status or raw client data. In particular, clients begin the identification stage from the same ImageNet-pretrained initialization and locally optimize the model using their observed local dataset and the LA objective defined below. For client $k$, the empirical class prior was
\begin{equation}
    \pi_{k,c}
    =
    \frac{n_{k,c}}
    {\sum_{j=1}^{C}n_{k,j}},
\end{equation}
where $n_{k,c}$ denotes the number of local samples carrying observed
label $c$. The class-specific LA was
\begin{equation}
    m_{k,c}
    =
    \beta\log(\pi_{k,c}+\epsilon),
\end{equation}
where $\beta$ controls the strength of the adjustment and $\epsilon$
prevents numerical instability. Let
$\mathbf{m}_k=[m_{k,1},\ldots,m_{k,C}]^\top$.
The Stage-I objective was
\begin{equation}
    \mathcal{L}_{\mathrm{LA}}
    =
    \operatorname{CE}
    \left(
        f_{\boldsymbol{\theta}_k}(x_i)
        +
        \mathbf{m}_k,
        \widetilde{y}_i
    \right).
\end{equation}
After local identification training, latent representations were extracted using deterministic transforms from the balanced loader. Let
$
    \mathbf{z}_i
    =
    h_{\boldsymbol{\theta}_k}(x_i)
    \in \mathbb{R}^{d}
$
be the feature vector for sample $i$. For each locally observed class $c$, the feature matrix was
\begin{equation}
    \mathbf{Z}_{k,c}
    =
    \begin{bmatrix}
        \mathbf{z}_1^{\top} \\
        \mathbf{z}_2^{\top} \\
        \vdots \\
        \mathbf{z}_{n_{k,c}}^{\top}
    \end{bmatrix}
    \in \mathbb{R}^{n_{k,c} \times d}.
\end{equation}
Singular value decomposition (SVD) was applied to each observed class matrix with descending singular values:
$
    \mathbf{Z}_{k,c}
    =
    \mathbf{U}_{k,c}
    \boldsymbol{\Sigma}_{k,c}
    \mathbf{V}_{k,c}^{\top}.
$
The leading right singular vector $\mathbf{v}_{k,c}$ represents the dominant direction of class $c$. Pairwise class similarity was calculated as
\begin{equation}
    [\mathbf{S}_k]_{c,c'}
    =
    \left|
    \frac{
        \mathbf{v}_{k,c}^{\top}\mathbf{v}_{k,c'}
    }{
        \lVert \mathbf{v}_{k,c}\rVert_2
        \lVert \mathbf{v}_{k,c'}\rVert_2
        + \epsilon
    }
    \right|.
\end{equation}
Under reliable annotations, class-discriminative feature representations are expected to produce distinct dominant directions across diagnostic categories. Label corruption mixes samples from different underlying classes within the same observed category and introduces shared class components across category specific feature matrices. Consequently, the dominant directions tend to exhibit greater cross-class alignment. This observation motivates using off-diagonal spectral similarity as an indicator of annotation unreliability. Therefore, lower cross-class alignment was taken as evidence of more distinct class-specific feature directions, whereas increased overlap was treated as a marker of label unreliability as shown in Fig.~\ref{fig:spectral_client_identification}.

Each client was summarized by three statistics computed from the off-diagonal entries: the mean off-diagonal similarity, mean squared magnitude (energy), and maximum:
\begin{equation}
    \mu_k
    =
    \frac{1}{|\Omega_k|}
    \sum_{(c,c') \in \Omega_k}
    [\mathbf{S}_k]_{c,c'},
\end{equation}
\begin{equation}
    e_k
    =
    \frac{1}{|\Omega_k|}
    \sum_{(c,c') \in \Omega_k}
    [\mathbf{S}_k]_{c,c'}^2,
\end{equation}
\begin{equation}
    s^{\max}_k
    =
    \max_{(c,c') \in \Omega_k}
    [\mathbf{S}_k]_{c,c'}.
\end{equation}
where
$
    \Omega_k
    =
    \{(c,c') : c \neq c',\; c,c' \in \mathcal{C}_k\}.
$
A two-component GMM was fitted to the client descriptors. The component exhibiting lower cross-class similarity was designated as clean, producing the estimated sets
${\mathcal{K}}_{\mathrm{clean}}$ and 
${\mathcal{K}}_{\mathrm{noisy}}.$
No ground-truth client status was used during this partitioning.

\noindent
\subsubsection*{Class-specific representative and residual subspaces}

At each relabeling checkpoint, the clean reference model was used to extract features from clients in $\widehat{\mathcal{K}}_{\mathrm{clean}}$. For each locally observed class $c$ at client $k$, SVD was applied to the corresponding class-specific feature matrix, $
\mathbf{Z}_{k,c}
=
\mathbf{U}_{k,c}
\boldsymbol{\Sigma}_{k,c}
\mathbf{V}_{k,c}^{\top},
$
where the singular values
$\sigma_1 \geq \sigma_2 \geq \cdots$
were ordered in descending magnitude. An adaptive spectral cutoff $q_{k,c}$ was selected as the smallest integer satisfying
\begin{equation}
\sum_{j=1}^{q_{k,c}} \sigma_j
\geq
\kappa
\sum_{j=q_{k,c}+1}^{R_{k,c}} \sigma_j,
\end{equation}
where $R_{k,c}$ denotes the available spectral rank and $\kappa$ controls the relative spectral mass of the dominant and residual portions. Equivalently,
\begin{equation}
\sum_{j=1}^{q_{k,c}} \sigma_j
\geq
\frac{\kappa}{\kappa+1}
\sum_{j=1}^{R_{k,c}} \sigma_j.
\end{equation}
We used $\kappa=10$, corresponding to a cutoff at which the dominant portion contained at least approximately $90.9\%$ of the total singular-value mass. The leading right singular vector,
\begin{equation}
\mathbf{v}^{(r)}_{k,c}
=
\mathbf{v}_{k,c,1}
\in
\mathbb{R}^{d},
\end{equation}
defined the local representative direction. The residual portion began after the adaptive spectral cutoff $q_{k,c}$. The first $L$ right singular vectors after the cutoff were retained to construct the local residual basis,
\begin{equation}
\mathbf{V}^{(n)}_{k,c}
=
\left[
\mathbf{v}_{k,c,q_{k,c}+1},
\ldots,
\mathbf{v}_{k,c,q_{k,c}+L}
\right].
\end{equation}
where $L$ denotes the number of residual directions used for spectral reference construction. We used $L=12$ in all experiments. When fewer than $L$ residual directions were available for a local class-specific decomposition, all available residual directions were used. Because singular vectors are defined only up to sign and their orientations may vary across clients, the class-specific references were combined through weighted projector averaging rather than direct vector averaging. For class $c$, the representative projector was
\begin{equation}
\mathbf{P}^{(r)}_c
=
\frac{1}{W_c}
\sum_{k \in {\mathcal{K}}_{\mathrm{clean}}}
w_{k,c}
\mathbf{v}_{k,c}^{(r)}
\mathbf{v}_{k,c}^{(r)\top},
\end{equation}
and the residual projector was
\begin{equation}
\mathbf{P}^{(n)}_c
=
\frac{1}{W_c}
\sum_{k \in {\mathcal{K}}_{\mathrm{clean}}}
w_{k,c}
\mathbf{V}_{k,c}^{(n)}
\mathbf{V}_{k,c}^{(n)\top},
\end{equation}
with
\begin{equation}
W_c
=
\sum_{k \in {\mathcal{K}}_{\mathrm{clean}}}
w_{k,c},
\end{equation}
where $w_{k,c}$ is the number of samples belonging to class $c$ at client $k$.
The principal eigenvector of $\mathbf{P}^{(r)}_c$ defined the global representative direction $\overline{\mathbf{v}}^{(r)}_c$, whereas the leading $L$ eigenvectors of $\mathbf{P}^{(n)}_c$ defined the global residual basis $\overline{\mathbf{V}}^{(n)}_c$. We used the same $L=12$ residual directions to construct the global residual reference.

\subsubsection*{Spectral relabeling of noisy clients}

\begin{figure}[htbp]
    \centering
    \includegraphics[width=0.95\textwidth]{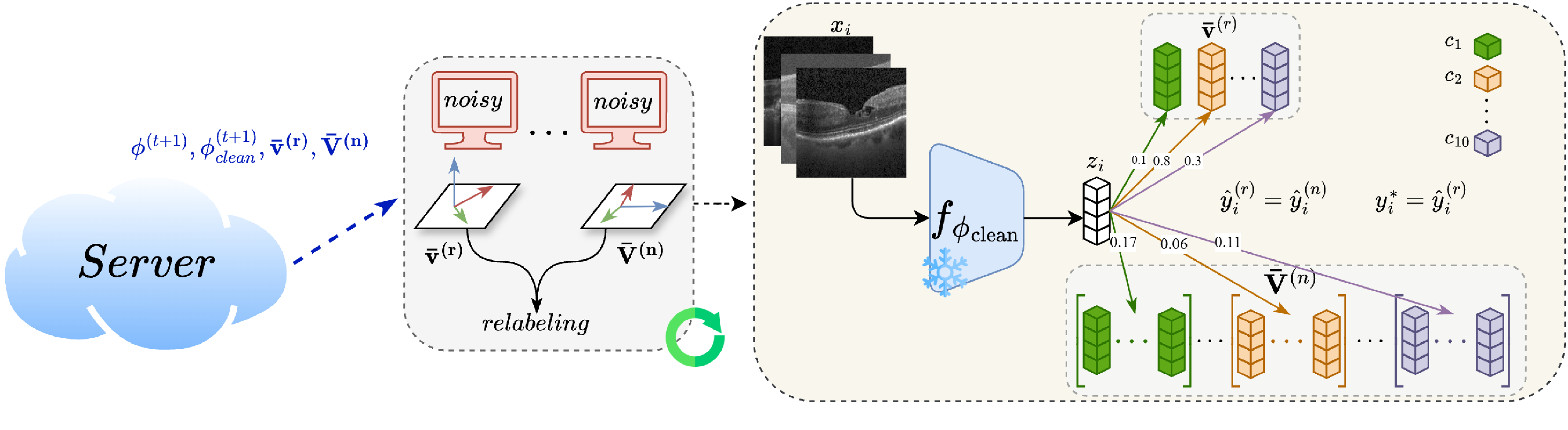}
    \caption{Spectral relabeling of predicted noisy clients using representative and
residual spectral references constructed from predicted clean clients.
A label change is accepted only when the representative- and
residual-subspace predictions agree.}
    \label{fig:relabeling}
\end{figure}

The clean reference model was used to extract features from each predicted noisy client (Fig. \ref{fig:relabeling}). 

\noindent For a feature vector $\mathbf{z}_i$, representative alignment with class $c$ was
\begin{equation}
    S_R(i,c)
    =
    \left|
        \overline{\mathbf{v}}^{(r)\top}_c
        \mathbf{z}_i
    \right|.
\end{equation}
where larger values indicate stronger alignment with the representative class direction. Residual-subspace projection was
\begin{equation}
    S_N(i,c)
    =
    \frac{1}{\sqrt{L}}
    \left\lVert
        \overline{\mathbf{V}}^{(n)\top}_c
        \mathbf{z}_i
    \right\rVert_2.
\end{equation}
The two spectral predictions were
\begin{equation}
    \widehat{y}_i^{(R)}
    =
    \arg\max_c S_R(i,c),
\end{equation}
\begin{equation}
    \widehat{y}_i^{(N)}
    =
    \arg\min_c S_N(i,c).
\end{equation}
The final hard target was defined as
\begin{equation}
    y_i^{\star}
    =
    \begin{cases}
        \widehat{y}_i^{(R)},
        &
        \widehat{y}_i^{(R)}
        =
        \widehat{y}_i^{(N)}, \\[4pt]
        \widetilde{y}_i,
        &
        \text{otherwise}.
    \end{cases}
\end{equation}
Thus, a candidate correction was accepted only when the representative-
and residual-subspace criteria agreed; otherwise, the observed
annotation was retained. This agreement criterion reduced the risk of
replacing uncertain clinical labels with unstable pseudo-labels. 

\subsubsection*{Local optimization}
Predicted clean clients continued to use the LA objective defined above, with class priors estimated from their observed local labels. For a predicted noisy client, spectral relabeling produced a corrected hard target $y_i^{\star}$. When the representative- and residual-subspace predictions agreed, $y_i^{\star}$ was replaced by the agreed spectral prediction; otherwise,
$y_i^{\star}=\widetilde{y}_i$.
Thus, every sample retained a valid hard training target, while label changes were restricted to samples supported by both spectral criteria. The current global model served as the teacher for KD. For temperature $\tau$, the teacher distribution for sample $x_i$ was
\begin{equation}
\mathbf{p}_i^{(T)}
=
\operatorname{softmax}
\left(
\frac{
f_{\boldsymbol{\theta}^{(t)}}(x_i)
}{\tau}
\right).
\end{equation}
For the noisy-client model $(k' \in \mathcal{K}_{\mathrm{noisy}}$), the local logits were first adjusted using the client-specific class prior,
\begin{equation}
\boldsymbol{\ell}_{k',i}
=
f_{\boldsymbol{\theta}^{(t)}_{k'}}(x_i)
+
\mathbf{m}_k.
\end{equation}
The corresponding student distribution was
\begin{equation}
\mathbf{p}_i^{(S)}
=
\operatorname{softmax}
\left(
\frac{
\boldsymbol{\ell}_{k',i}
}{\tau}
\right).
\end{equation}
The distillation loss was
\begin{equation}
\mathcal{L}_{\mathrm{KD}}
=
D_{\mathrm{KL}}
\left(
\mathbf{p}_i^{(T)}
\,\|\,
\mathbf{p}_i^{(S)}
\right)
\end{equation}
whereas supervision from the spectrally corrected hard target was
\begin{equation}
\mathcal{L}_{\mathrm{hard}}
=
\operatorname{CE}
\left(
\boldsymbol{\ell}_{k',i},
y_i^{\star}
\right).
\end{equation}
The noisy-client objective combines the two sources of supervision:
\begin{equation}
\mathcal{L}_{\mathrm{noisy}}^{(t)}
=
\lambda
\mathcal{L}_{\mathrm{KD}}
+
\left(1-\lambda\right)
\mathcal{L}_{\mathrm{hard}},
\end{equation}
where $\lambda$ corresponds to the estimated fraction of clean clients
in the federation and controls the relative contribution of global-model
distillation and corrected hard-label supervision:
\begin{equation}
    \lambda
    =
    \frac{
        \left|
        {\mathcal{K}}_{\mathrm{clean}}
        \right|
    }{K}.
\end{equation} Before the first spectral relabeling checkpoint, no corrected labels were yet available; therefore, $y_i^{\star}=\widetilde{y}_i$, and noisy clients were trained using their observed labels together with KD from the current global model. At subsequent relabeling checkpoints, accepted spectral corrections were incorporated into $y_i^{\star}$ and used during local optimization.

\subsubsection*{Distance-aware aggregation (DaAgg)}

After local optimization, all clients remained eligible to contribute to the global model. Let
\begin{equation}
    a_k
    =
    \frac{N_k}{\sum_j N_j}
\end{equation}
be the conventional sample-size weight. For each predicted noisy client, its parameter-space distance to the nearest predicted clean client was calculated as
\begin{equation}
    d_k
    =
    \min_{j \in \widehat{\mathcal{K}}_{\mathrm{clean}}}
    \sum_{\ell}
    \left\lVert
        \boldsymbol{\theta}_{k,\ell}
        -
        \boldsymbol{\theta}_{j,\ell}
    \right\rVert_2.
\end{equation}
where $\ell$ indexes floating-point parameter tensors. For predicted clean clients, $d_k=0$. Distances were normalized as
\begin{equation}
    \overline{d}_k
    =
    \frac{d_k}{\max_j d_j+\epsilon}.
\end{equation}
The final aggregation weight was
\begin{equation}
    \omega_k
    =
    \frac{
        a_k\exp(-\overline{d}_k)
    }{
        \sum_j a_j\exp(-\overline{d}_j)
    },
\end{equation}
and the global model was updated by
\begin{equation}
    \boldsymbol{\theta}^{(t+1)}
    =
    \sum_{k=1}^{K}
    \omega_k\boldsymbol{\theta}^{(t)}_k.
\end{equation}
DaAgg therefore reduced the contribution of noisy-client models that deviated substantially from all clean-client models without removing them entirely. In parallel, the predicted clean-client models were aggregated using sample-size-weighted averaging to update the clean reference model used at the next relabeling checkpoint.





\subsection*{Network architectures}
We used lightweight ImageNet-pretrained convolutional neural networks to limit local computation, memory use, and communication cost. SqueezeNet1.1 was used for Kermany, ShuffleNetV2 for UIC, and RegNetY-400MF for WF. Compact architectures are relevant to FL because each participating client repeatedly stores the model, performs local optimization, and transmits model updates \cite{bonawitz2019towards}. SqueezeNet reduces parameter count through Fire modules that combine $1\times1$ and $3\times3$ convolutions \cite{squeezenet}; ShuffleNetV2 was designed for efficient execution with attention to memory-access cost and practical latency \cite{Shufflenet}; and RegNetY-400MF provides a regular network design within a low computational budget \cite{regnet}. For each dataset, the ImageNet classifier was replaced by a linear layer matching the number of target classes, and the backbone and classifier were fine-tuned jointly. Within each dataset, all FL methods used the same architecture and initialization.

\subsection*{Training protocol and reproducibility}

Models were initialized using ImageNet-pretrained parameters. Random seeds were set consistently for NumPy, PyTorch, client partitioning, label corruption, data-loader workers, and batch generators. Within a given experimental condition, the initial model state was generated once and shared across FL methods. All experiments were repeated using three independent random seeds.

Local optimization used Adam or the optimizer prescribed by the corresponding baseline. OCT-FedSIR used Adam with weight decay $5\times10^{-4}$ during its noise-aware training stage. Training was performed for 100 communication rounds. In each round, each client received the current global model, performed one local epoch with a batch size of 128, and returned its updated state to the server. Each dataset-specific backbone was evaluated at every symmetric and asymmetric noise ratio and at each Dirichlet heterogeneity level.

For all three ophthalmic datasets, images were resized to
$128\times128$ pixels and converted to tensors. During training,
random horizontal flipping was applied, followed by channel-wise
normalization using mean
$0.5$ and standard deviation $0.5$.
Evaluation used the same resizing and normalization without random augmentation.
\subsection*{Evaluation and statistical analysis}
The global model was evaluated on the held-out test set after each communication round to characterize the training trajectory. Unless otherwise stated, all classification results reported in the tables, figures, and text correspond to the global model obtained after the final communication round. Test-set performance was not used for checkpoint selection, client identification, spectral relabeling, local optimization, or any other aspect of model training. Classification performance was summarized using accuracy and macro-F1 score.

Client-identification performance was evaluated against the known simulated client status $\gamma_k$. We reported overall identification accuracy, clean-client precision, clean-client recall, clean-client F1 score, noisy-client recall, and the numbers of clean clients incorrectly classified as noisy and noisy clients incorrectly classified as clean. Ground-truth client status was used only for retrospective evaluation and was not provided to the GMM during client identification.

Sample-level relabeling performance was evaluated by comparing the original reference label $y_i$, the simulated corrupted label $\widetilde{y}_i$, and the post-correction label $y_i^{\star}$. Let $N$ denote the number of samples evaluated during the relabeling procedure. The edit rate was defined as

\begin{equation}
    \text{Edit rate}
    =
    \frac{
        \#\{i:y_i^{\star}\neq\widetilde{y}_i\}
    }{N}.
\end{equation}
The recovery rate quantifies the proportion of intentionally corrupted annotations that were restored to their original reference labels:

\begin{equation}
    \text{Recovery rate}
    =
    \frac{
        \#\{
        i:
        \widetilde{y}_i\neq y_i
        \land
        y_i^{\star}=y_i
        \}
    }{
        \#\{
        i:
        \widetilde{y}_i\neq y_i
        \}
    }.
\end{equation}
Correction precision quantifies the proportion of accepted label changes that resulted in the original reference label:

\begin{equation}
    \text{Correction precision}
    =
    \frac{
        \#\{
        i:
        y_i^{\star}\neq\widetilde{y}_i
        \land
        y_i^{\star}=y_i
        \}
    }{
        \#\{
        i:
        y_i^{\star}\neq\widetilde{y}_i
        \}
    }.
\end{equation}
The false-correction rate quantifies the proportion of annotations that were originally correct but became incorrect after spectral relabeling:

\begin{equation}
    \text{False-correction rate}
    =
    \frac{
        \#\{
        i:
        \widetilde{y}_i=y_i
        \land
        y_i^{\star}\neq y_i
        \}
    }{
        \#\{
        i:
        \widetilde{y}_i=y_i
        \}
    }.
\end{equation}
The residual label-noise rate after spectral relabeling was defined as

\begin{equation}
    \text{Residual noise rate}
    =
    \frac{
        \#\{
        i:
        y_i^{\star}\neq y_i
        \}
    }{N}.
\end{equation}

All experimental conditions were repeated using three independent random seeds. For each seed, client partitioning, clean/noisy client assignment, label corruption, and model initialization were generated independently. Within a given seed and experimental condition, however, the same client partition, clean/noisy client assignment, corrupted sample indices, corrupted labels, and initial model state were shared across all competing federated learning methods to enable paired and directly comparable evaluations.

Classification, client-identification, and relabeling metrics were computed independently for each seed. Unless otherwise stated, numerical results are reported as the mean $\pm$ standard deviation across the three independent experimental runs. For each run, the evaluation pipeline recorded per-round predictions, per-sample losses, aggregate and class-wise classification metrics, client-identification outcomes, and sample-level relabeling outcomes.
\subsection*{Use of generative artificial intelligence}
ChatGPT (OpenAI) was used solely for language editing, including grammatical correction and rephrasing of manuscript text. It was not used to generate or analyze experimental data, perform statistical analyses, or determine scientific conclusions. All revised text was reviewed and approved by the authors.

{\small
\bibliographystyle{IEEEtran}
\bibliography{references}

@article{kermany,
  title={Identifying medical diagnoses and treatable diseases by image-based deep learning},
  author={Kermany, Daniel S and Goldbaum, Michael and Cai, Wenjia and Valentim, Carolina CS and Liang, Huiying and Baxter, Sally L and McKeown, Alex and Yang, Ge and Wu, Xiaokang and Yan, Fangbing and others},
  journal={cell},
  volume={172},
  number={5},
  pages={1122--1131},
  year={2018},
  publisher={Elsevier}
}

@article{Fauw,
  title={Clinically applicable deep learning for diagnosis and referral in retinal disease},
  author={De Fauw, Jeffrey and Ledsam, Joseph R and Romera-Paredes, Bernardino and Nikolov, Stanislav and Tomasev, Nenad and Blackwell, Sam and Askham, Harry and Glorot, Xavier and O’Donoghue, Brendan and Visentin, Daniel and others},
  journal={Nature medicine},
  volume={24},
  number={9},
  pages={1342--1350},
  year={2018},
  publisher={Nature Publishing Group US New York}
}

@inproceedings{macmahan,
  title={Communication-efficient learning of deep networks from decentralized data},
  author={McMahan, Brendan and Moore, Eider and Ramage, Daniel and Hampson, Seth and y Arcas, Blaise Aguera},
  booktitle={Artificial intelligence and statistics},
  pages={1273--1282},
  year={2017},
  organization={Pmlr}
}

@article{lo2021federated,
  title={Federated learning for microvasculature segmentation and diabetic retinopathy classification of OCT data},
  author={Lo, Julian and Timothy, T Yu and Ma, Da and Zang, Pengxiao and Owen, Julia P and Zhang, Qinqin and Wang, Ruikang K and Beg, Mirza Faisal and Lee, Aaron Y and Jia, Yali and others},
  journal={Ophthalmology Science},
  volume={1},
  number={4},
  pages={100069},
  year={2021},
  publisher={Elsevier}
}

@article{lu2022federated,
  title={Federated learning for multicenter collaboration in ophthalmology: improving classification performance in retinopathy of prematurity},
  author={Lu, Charles and Hanif, Adam and Singh, Praveer and Chang, Ken and Coyner, Aaron S and Brown, James M and Ostmo, Susan and Chan, Robison V Paul and Rubin, Daniel and Chiang, Michael F and others},
  journal={Ophthalmology Retina},
  volume={6},
  number={8},
  pages={657--663},
  year={2022},
  publisher={Elsevier}
}

@article{gholami2023federated,
  title={Federated learning for diagnosis of age-related macular degeneration},
  author={Gholami, Sina and Lim, Jennifer I and Leng, Theodore and Ong, Sally Shin Yee and Thompson, Atalie Carina and Alam, Minhaj Nur},
  journal={Frontiers in Medicine},
  volume={10},
  pages={1259017},
  year={2023},
  publisher={Frontiers Media SA}
}

@article{ran2024developing,
  title={Developing a privacy-preserving deep learning model for glaucoma detection: a multicentre study with federated learning},
  author={Ran, An Ran and Wang, Xi and Chan, Poemen P and Wong, Mandy OM and Yuen, Hunter and Lam, Nai Man and Chan, Noel CY and Yip, Wilson WK and Young, Alvin L and Yung, Hon-Wah and others},
  journal={British Journal of Ophthalmology},
  volume={108},
  number={8},
  pages={1114--1123},
  year={2024},
  publisher={BMJ Publishing Group Ltd}
}

@article{gholami-distributed,
  title={Distributed training of foundation models for ophthalmic diagnosis},
  author={Gholami, Sina and Jannat, Fatema-E and Thompson, Atalie Carina and Ong, Sally Shin Yee and Lim, Jennifer I and Leng, Theodore and Tabkhivayghan, Hamed and Alam, Minhaj Nur},
  journal={Communications Engineering},
  volume={4},
  number={1},
  pages={6},
  year={2025},
  publisher={Nature Publishing Group UK London}
}

@article{fbnll,
  title={{FB-NLL}: A Feature-Based Approach to Tackle Noisy Labels in Personalized Federated Learning},
  author={Ali, A. and Arafa, A.},
note={Available online: arXiv:2604.19729}
}

@article{miladinovic2024evaluating,
  title={Evaluating deep learning models for classifying OCT images with limited data and noisy labels},
  author={Miladinovi{\'c}, Aleksandar and Biscontin, Alessandro and Aj{\v{c}}evi{\'c}, Milo{\v{s}} and Kresevic, Simone and Accardo, Agostino and Marangoni, Dario and Tognetto, Daniele and Inferrera, Leandro},
  journal={Scientific Reports},
  volume={14},
  number={1},
  pages={30321},
  year={2024},
  publisher={Nature Publishing Group UK London}
}

@article{nabil2025federated,
  title={Federated Learning for Multi-Disease Ophthalmic Diagnostics using Optical Coherence Tomography Angiography (OCTA)},
  author={Nabil, Ahammed Sakir and Gholami, Sina and Leng, Theodore and Lim, Jennifer I and Alam, Minhaj Nur},
  journal={Ophthalmology Science},
  pages={101030},
  year={2025},
  publisher={Elsevier}
}

@inproceedings{lau2025fedted,
  title={Fedted: Federated learning for robust thyroid eye disease detection with masked autoencoders},
  author={Lau, Wai Tak and McCarthy, Angela and Tian, Ye and Nielsen, Christopher and Chelliah, Rashmi and Gholami, Sina and Kossler, Andrea and Alam, Minhaj and Glass, Lora Dagi and Thakoor, Kaveri A},
  booktitle={2025 IEEE EMBS International Conference on Biomedical and Health Informatics (BHI)},
  pages={1--7},
  year={2025},
  organization={IEEE}
}

@article{yonamine2024comparison,
  title={Comparison of diagnosis codes to clinical notes in classifying patients with diabetic retinopathy},
  author={Yonamine, Sean and Ma, Chu Jian and Alabi, Rolake O and Kaidonis, Georgia and Chan, Lawrence and Borkar, Durga and Stein, Joshua D and Arnold, Benjamin F and Sun, Catherine Q},
  journal={Ophthalmology Science},
  volume={4},
  number={6},
  pages={100564},
  year={2024},
  publisher={Elsevier}
}

@article{ju2022improving,
  title={Improving medical images classification with label noise using dual-uncertainty estimation},
  author={Ju, Lie and Wang, Xin and Wang, Lin and Mahapatra, Dwarikanath and Zhao, Xin and Zhou, Quan and Liu, Tongliang and Ge, Zongyuan},
  journal={IEEE transactions on medical imaging},
  volume={41},
  number={6},
  pages={1533--1546},
  year={2022},
  publisher={IEEE}
}

@article{campbell2022artificial,
  title={Artificial intelligence for retinopathy of prematurity: validation of a vascular severity scale against international expert diagnosis},
  author={Campbell, J Peter and Chiang, Michael F and Chen, Jimmy S and Moshfeghi, Darius M and Nudleman, Eric and Ruambivoonsuk, Paisan and Cherwek, Hunter and Cheung, Carol Y and Singh, Praveer and Kalpathy-Cramer, Jayashree and others},
  journal={Ophthalmology},
  volume={129},
  number={7},
  pages={e69--e76},
  year={2022},
  publisher={Elsevier}
}

@article{lin2025efficiency,
  title={Efficiency and safety of automated label cleaning on multimodal retinal images},
  author={Lin, Tian and Wang, Meng and Lin, Aidi and Mai, Xiaoting and Liang, Huiyu and Tham, Yih-Chung and Chen, Haoyu},
  journal={npj Digital Medicine},
  volume={8},
  number={1},
  pages={10},
  year={2025},
  publisher={Nature Publishing Group UK London}
}

@article{rofl,
  title={Robust federated learning with noisy labels},
  author={Yang, S. and Park, H. and Byun, J. and Kim, C.},
  journal={IEEE Intelligent Systems},
  volume={37},
  number={2},
  pages={35--43},
  year={April  2022},
  publisher={IEEE}
}

@inproceedings{rhfl,
  title={Robust federated learning with noisy and heterogeneous clients},
  author={Fang, Xiuwen and Ye, Mang},
  booktitle={2022 IEEE/CVF Conference on Computer Vision and Pattern Recognition (CVPR)},
  pages={10062--10071},
  year={2022},
  organization={IEEE}
}

@inproceedings{fedcorr,
  title={Fedcorr: Multi-stage federated learning for label noise correction},
  author={Xu, Jingyi and Chen, Zihan and Quek, Tony QS and Chong, Kai Fong Ernest},
  booktitle={2022 IEEE/CVF Conference on Computer Vision and Pattern Recognition (CVPR)},
  pages={10174--10183},
  year={2022},
  organization={IEEE}
}

@inproceedings{fedned,
  title={Federated learning with extremely noisy clients via negative distillation},
  author={Lu, Yang and Chen, Lin and Zhang, Yonggang and Zhang, Yiliang and Han, Bo and Cheung, Yiu-ming and Wang, Hanzi},
  booktitle={Proceedings of the AAAI conference on artificial intelligence},
  volume={38},
  number={13},
  pages={14184--14192},
  year={2024}
}

@article{fednoro,
  title={FedNoRo: Towards noise-robust federated learning by addressing class imbalance and label noise heterogeneity},
  author={Wu, Nannan and Yu, Li and Jiang, Xuefeng and Cheng, Kwang-Ting and Yan, Zengqiang},
  journal={arXiv preprint arXiv:2305.05230},
  year={2023}
}

@inproceedings{fedelc,
  title={Tackling noisy clients in federated learning with end-to-end label correction},
  author={Jiang, Xuefeng and Sun, Sheng and Li, Jia and Xue, Jingjing and Li, Runhan and Wu, Zhiyuan and Xu, Gang and Wang, Yuwei and Liu, Min},
  booktitle={Proceedings of the 33rd ACM international conference on information and knowledge management},
  pages={1015--1026},
  year={2024}
}

@inproceedings{fedsir,
  title={FedSIR: Spectral Client Identification and Relabeling for Federated Learning with Noisy Labels},
  author={Gholami, Sina and Ali, Abudlmoneam and Haghighi, Tania and Arafa, Ahmed and Alam, Minhaj Nur},
  booktitle={Proceedings of the IEEE/CVF Conference on Computer Vision and Pattern Recognition},
  pages={3340--3348},
  year={2026}
}

@article{bonawitz2019towards,
  title={Towards federated learning at scale: System design},
  author={Bonawitz, Keith and Eichner, Hubert and Grieskamp, Wolfgang and Huba, Dzmitry and Ingerman, Alex and Ivanov, Vladimir and Kiddon, Chloe and Kone{\v{c}}n{\`y}, Jakub and Mazzocchi, Stefano and McMahan, Brendan and others},
  journal={Proceedings of machine learning and systems},
  volume={1},
  pages={374--388},
  year={2019}
}

@article{squeezenet,
  title={SqueezeNet: AlexNet-level accuracy with 50x fewer parameters and< 0.5 MB model size},
  author={Iandola, Forrest N and Han, Song and Moskewicz, Matthew W and Ashraf, Khalid and Dally, William J and Keutzer, Kurt},
  journal={arXiv preprint arXiv:1602.07360},
  year={2016}
}

@inproceedings{Shufflenet,
  title={Shufflenet v2: Practical guidelines for efficient cnn architecture design},
  author={Ma, Ningning and Zhang, Xiangyu and Zheng, Hai-Tao and Sun, Jian},
  booktitle={Proceedings of the European conference on computer vision (ECCV)},
  pages={116--131},
  year={2018}
}

@inproceedings{fedlsr,
  title={Towards federated learning against noisy labels via local self-regularization},
  author={Jiang, X. and Sun, S. and Wang, Y. and Liu, M.},
  booktitle={Proc. ACM CIKM},
  year={October 2022}
}

@article{fedprox,
  title={Federated optimization in heterogeneous networks},
  author={Li, T. and Sahu, A. K. and Zaheer, M. and Sanjabi, M. and Talwalkar, A. and Smith, V.},
  journal={Proc. Mach. Learn. Syst.},
  volume={2},
  pages={429--450},
  year={March 2020},

}

@article{regnet,
  title={Designing network design spaces},
  author={Radosavovic, Ilija and Kosaraju, Raj Prateek and Girshick, Ross and He, Kaiming and Doll{\'a}r, Piotr},
  journal={arXiv preprint arXiv:2003.13678},
  year={2020}
}

@inproceedings{fedsim,
  title={FedSim: foundational federated multi-task learning for ophthalmic diagnostics},
  author={Gholami, Sina and Ali, Abdulmoneam and Kamanda, Hindolo and Haghighi, Tania and Ong, Sally Shin Yee and Lim, Jennifer I and Leng, Theodore and Arafa, Ahmed and Alam, Minhaj Nur},
  booktitle={Ophthalmic Technologies XXXVI},
  volume={13831},
  pages={140--149},
  year={2026},
  organization={SPIE}
}

@article{haghighi2025compact,
  title={Compact vision language models enable efficient and interpretable optical coherence tomography through layer-specific multimodal learning},
  author={Haghighi, Tania and Gholami, Sina and Sokol, Jared Todd and Biswas, Aayush and Lim, Jennifer I and Leng, Theodore and Thompson, Atalie C and Tabkhi, Hamed and Alam, Minhaj Nur},
  journal={Communications Medicine},
  volume={6},
  number={1},
  pages={32},
  year={2025},
  publisher={Nature Publishing Group UK London}
}

@article{lee2017deep,
  title={Deep learning is effective for classifying normal versus age-related macular degeneration OCT images},
  author={Lee, Cecilia S and Baughman, Doug M and Lee, Aaron Y},
  journal={Ophthalmology Retina},
  volume={1},
  number={4},
  pages={322--327},
  year={2017},
  publisher={Elsevier}
}

@article{schlegl2018,
  title={Fully automated detection and quantification of macular fluid in OCT using deep learning},
  author={Schlegl, Thomas and Waldstein, Sebastian M and Bogunovic, Hrvoje and Endstra{\ss}er, Franz and Sadeghipour, Amir and Philip, Ana-Maria and Podkowinski, Dominika and Gerendas, Bianca S and Langs, Georg and Schmidt-Erfurth, Ursula},
  journal={Ophthalmology},
  volume={125},
  number={4},
  pages={549--558},
  year={2018},
  publisher={Elsevier}
}

@article{siraz2025multi,
  title={Multi-class classification of central and non-central geographic atrophy using Optical Coherence Tomography},
  author={Siraz, Sadia and Kamanda, Hindolo and Gholami, Sina and Nabil, Ahammed Sakir and Yee Ong, Sally Shin and Alam, Minhaj Nur},
  journal={medRxiv},
  pages={2025--05},
  year={2025},
  publisher={Cold Spring Harbor Laboratory Press}
}

@article{jannat2026multi,
  title={Multi-OCT-SelfNet: Integrating self-supervised learning with multi-source data fusion for enhanced multi-class retinal disease classification},
  author={Jannat, Fatema E and Gholami, Sina and Lim, Jennifer I and Leng, Theodore and Alam, Minhaj Nur and Tabkhi, Hamed},
  journal={Frontiers in Systems Biology},
  volume={6},
  pages={1717398},
  year={2026}
}

@article{karimi2020noisy,
  title={Deep learning with noisy labels: Exploring techniques and remedies in medical image analysis},
  author={Karimi, Davood and Dou, Haoran and Warfield, Simon K and Gholipour, Ali},
  journal={Medical image analysis},
  volume={65},
  pages={101759},
  year={2020},
  publisher={Elsevier}
}

@article{shi2024labelnoise,
  title={A survey of label-noise deep learning for medical image analysis},
  author={Shi, Jialin and Zhang, Kailai and Guo, Chenyi and Yang, Youquan and Xu, Yali and Wu, Ji},
  journal={Medical image analysis},
  volume={95},
  pages={103166},
  year={2024},
  publisher={Elsevier}
}

@misc{OCT-FedSIR,
  author       = {QIAI UNCC},
  title        = {Toward Trustworthy Federated Ophthalmic Learning under Annotation Noise hosted on GitHub},
  year         = {2026},
  howpublished = {\url{https://github.com/QIAIUNCC/OCT-FedSIR-Toward-Trustworthy Federated-Ophthalmic-Learning-under-Annotation-Noise}},
  note         = {Accessed: 2026-09-05}
}
}
\section*{Acknowledgment} 
This study is supported by NEI R15EY035804, R21EY035271, 1R01EY037828-01 (MNA), NC Diabetes Research Center P30DK124723 (SSYO), Research to Prevent Blindness (TL) and NIH Grant P30EY026877 (TL). 

\section*{Author information}
\subsection*{Authors and affiliations}
Department of Electrical and Computer Engineering, University of North Carolina at Charlotte, Charlotte, NC, USA\\
Sina Gholami, Abdulmoneam Ali, Tania Haghighi, Rashadul H. Badhon, Behafarin Emam, Ahmed Arafa, \& Minhaj Nur Alam \\
Department of Ophthalmology, Wake Forest School of Medicine, Winston-Salem, NC, USA \\
Sally S.Y. Ong \&  Atalie C. Thompson \\
Byers Eye Institute at Stanford, Stanford University School of Medicine, Stanford, CA, USA\\
Theodore Leng\\
Department of Ophthalmology and Visual Sciences, University of Illinois Chicago, Chicago, IL, USA\\
Jennifer I. Lim

\affil[1]{Department of Electrical and Computer Engineering, University of North Carolina at Charlotte, Charlotte, NC, USA}
\affil[2]{Department of Ophthalmology, Wake Forest School of Medicine, Winston-Salem, NC, USA}
\affil[3]{Department of Ophthalmology and Visual Sciences, University of Illinois Chicago, Chicago, IL, USA}
\affil[4]{Byers Eye Institute at Stanford, Stanford University School of Medicine, Stanford, CA, USA}

\subsection*{Contributions}
Study conception and design: Sina Gholami, Abdulmoneam Ali, Tania Haghighi, Ahmed Arafa and Minhaj Nur Alam; Data collection and analysis: Sina Gholami, Rashadul H. Badhon, Sally S. Y. Ong, Atalie C. Thompson, Jennifer I. Lim, and Minhaj Nur Alam; Data visualization: Sina Gholami, Rashadul H. Badhon and Minhaj Nur Alam; Interpretation of results: Sina Gholami, Abdulmoneam Ali, Tania Haghighi, Behafarin Emam, Sally S. Y. Ong, Atalie C. Thompson, Theodore Leng, Jennifer I. Lim and Minhaj Nur Alam; Manuscript preparation: Sina Gholami, Abdulmoneam Ali, Tania Haghighi, Rashadul H. Badhon, Behafarin Emam and Minhaj Nur Alam; Manuscript review: Sina Gholami, Abdulmoneam Ali, Tania Haghighi, Rashadul H. Badhon, Behafarin Emam, Sally S. Y. Ong, Atalie C. Thompson, Theodore Leng, Ahmed Arafa, Jennifer I. Lim and Minhaj Nur Alam; Supervision: Minhaj Nur Alam; Funding acquisition: Minhaj Nur Alam; Project administration: Sina Gholami and Minhaj Nur Alam. 

\subsection*{Corresponding author}
Correspondence to \href{malam8@charlotte.edu}{Minhaj Nur Alam}.
\section*{Ethics declarations}
The UIC and WF datasets were collected under protocols approved by the respective Institutional Review Boards (IRBs) and in accordance with the ethical principles outlined in the Declaration of Helsinki.

\subsection*{Competing interests}
The authors declare no competing interests.

\subsection*{Data availability}
The Kermany dataset used in this study is publicly available. The private datasets from the University of Illinois at Chicago (UIC) and Wake Forest (WF) can be requested from the corresponding author, subject to reasonable conditions.

\subsection*{Code availability}
The code that supports the findings of this study is available on GitHub\cite{OCT-FedSIR}.
\newpage
\section*{Supplementary material}

\setcounter{figure}{0}
\renewcommand{\thefigure}{S\arabic{figure}}

\setcounter{table}{0}
\renewcommand{\thetable}{S\arabic{table}}

\begin{table*}[t]
\centering
\caption{Classification accuracy (\%) on Kermany under symmetric label noise.
Values are mean $\pm$ standard deviation across three independent seeds.
Bold and underline indicate the best- and second-best-performing noisy-label
FL methods, respectively. Pruning and all-clean FedAvg are shown as references.
}
\label{tab:kermany_sym}
\begin{adjustbox}{width=\textwidth}
\begin{tabular}{c c l c c c c c c c}
\toprule
\multicolumn{3}{c}{} & \multicolumn{7}{c}{Symmetric Noise} \\
\cmidrule(lr){4-10}
$\alpha$ & \# Clean Clients & Method 
& 30\% & 40\% & 50\% & 60\% & 70\% & 80\% & 90\% \\
\midrule

\multirow{10}{*}{2.0} & \multirow{10}{*}{3}
& FedAvg \cite{macmahan}   & 95.20$\pm$1.25 & 94.90$\pm$1.35 & 92.90$\pm$1.60 & 89.40$\pm$1.80 & 77.90$\pm$2.65 & 22.00$\pm$2.50 & 20.60$\pm$2.55 \\
& & FedProx \cite{fedprox} & 95.70$\pm$1.25 & 93.40$\pm$1.50 & 92.80$\pm$1.60 & 93.20$\pm$1.50 & 74.00$\pm$2.80 & 32.40$\pm$2.85 & 19.70$\pm$2.45 \\
& & RoFL \cite{rofl}       & 98.70$\pm$0.75 & 98.10$\pm$0.80 & 97.70$\pm$0.90 & \textbf{97.80$\pm$0.95} & 91.30$\pm$1.80 & \underline{92.10$\pm$1.75} & \underline{83.50$\pm$2.20} \\
& & RHFL \cite{rhfl}       & 78.07$\pm$3.88 & 76.86$\pm$4.06 & 69.40$\pm$4.30 & 60.79$\pm$4.16 & 41.52$\pm$5.37 & 30.85$\pm$6.16 & 28.97$\pm$4.81 \\
& & FedLSR \cite{fedlsr}   & 79.30$\pm$2.45 & 74.30$\pm$2.65 & 71.40$\pm$2.90 & 49.50$\pm$3.05 & 51.90$\pm$3.05 & 43.40$\pm$3.20 & 27.70$\pm$2.85 \\
& & FedCorr \cite{fedcorr} & \underline{98.80$\pm$0.65} & \underline{98.60$\pm$0.70} & \underline{98.00$\pm$0.90} & \textbf{97.80$\pm$0.90} & \underline{91.40$\pm$1.75} & 83.90$\pm$2.25 & 58.50$\pm$3.05 \\
& & FedNed \cite{fedned}   & 95.10$\pm$1.35 & 93.90$\pm$1.50 & 94.20$\pm$1.45 & 92.10$\pm$1.75 & 72.70$\pm$2.65 & 23.00$\pm$2.50 & 22.90$\pm$2.60 \\
& & FedELC \cite{fedelc}   & 76.90$\pm$2.50 & 70.40$\pm$2.70 & 62.70$\pm$2.90 & 50.60$\pm$3.35 & 49.10$\pm$3.30 & 27.80$\pm$2.70 & 21.00$\pm$2.40 \\
& & FedNoRo \cite{fednoro} & 97.80$\pm$0.85 & 97.00$\pm$1.05 & 96.80$\pm$1.10 & 78.30$\pm$2.55 & 29.20$\pm$2.90 & 35.00$\pm$2.85 & 25.00$\pm$2.70 \\
& & \textbf{OCT-FedSIR (Ours)} & \textbf{99.40$\pm$0.45} & \textbf{99.40$\pm$0.45} & \textbf{99.50$\pm$0.45} & \underline{95.60$\pm$1.30} & \textbf{95.10$\pm$1.35} & \textbf{96.00$\pm$1.20} & \textbf{93.60$\pm$1.60} \\
\cmidrule(lr){2-10}
& 3 & Pruning & \multicolumn{7}{c}{86.40 $\pm$ 0.95} \\
& 30 & FedAvg & \multicolumn{7}{c}{99.60 $\pm$ 0.35} \\   
\midrule
 
\multirow{10}{*}{0.5} & \multirow{10}{*}{3}
& FedAvg \cite{macmahan}   & 95.40$\pm$1.30 & 96.20$\pm$1.15 & 94.80$\pm$1.35 & 94.60$\pm$1.45 & 85.00$\pm$2.25 & 24.60$\pm$2.70 & 22.20$\pm$2.55 \\
& & FedProx \cite{fedprox} & 95.50$\pm$1.20 & 95.90$\pm$1.15 & 96.30$\pm$1.10 & 93.90$\pm$1.50 & 85.20$\pm$2.20 & 31.30$\pm$2.80 & 21.00$\pm$2.50 \\
& & RoFL \cite{rofl}       & 97.70$\pm$0.90 & 97.50$\pm$0.95 & 95.50$\pm$1.30 & 89.40$\pm$1.90 & 82.40$\pm$2.35 & 72.10$\pm$2.75 & 71.00$\pm$2.65 \\
& & RHFL \cite{rhfl}       & 62.08$\pm$6.35 & 60.36$\pm$5.93 & 54.22$\pm$5.58 & 48.40$\pm$5.31 & 33.93$\pm$5.08 & 28.78$\pm$5.45 & 27.57$\pm$3.82 \\
& & FedLSR \cite{fedlsr}   & 95.30$\pm$1.30 & 95.60$\pm$1.30 & 94.20$\pm$1.40 & 87.90$\pm$2.00 & 74.50$\pm$2.55 & 49.30$\pm$3.05 & 28.80$\pm$2.85 \\
& & FedCorr \cite{fedcorr} & \underline{98.60$\pm$0.70} & \underline{98.10$\pm$0.80} & 97.20$\pm$1.00 & \underline{96.70$\pm$1.00} & \underline{97.10$\pm$0.95} & \underline{95.70$\pm$1.20} & \underline{97.50$\pm$0.90} \\
& & FedNed \cite{fedned}   & 96.60$\pm$1.00 & 95.60$\pm$1.20 & 95.50$\pm$1.20 & 95.00$\pm$1.25 & 83.20$\pm$2.25 & 27.70$\pm$2.95 & 20.60$\pm$2.45 \\
& & FedELC \cite{fedelc}   & 83.30$\pm$2.40 & 78.90$\pm$2.55 & 68.70$\pm$2.80 & 61.00$\pm$3.15 & 56.60$\pm$3.15 & 49.50$\pm$3.20 & 51.20$\pm$2.90 \\
& & FedNoRo \cite{fednoro} & 97.80$\pm$0.85 & 97.80$\pm$0.95 & \underline{98.00$\pm$0.80} & 96.60$\pm$1.10 & 30.50$\pm$2.75 & 29.20$\pm$2.75 & 32.00$\pm$2.75 \\
& & \textbf{OCT-FedSIR (Ours)} & \textbf{99.50$\pm$0.45} & \textbf{99.70$\pm$0.35} & \textbf{99.00$\pm$0.60} & \textbf{97.90$\pm$0.85} & \textbf{97.30$\pm$1.00} & \textbf{97.70$\pm$0.90} & \textbf{98.00$\pm$0.90} \\

\cmidrule(lr){2-10}
& 3 & Pruning & \multicolumn{7}{c}{86.90 $\pm$ 1.05} \\
& 30 & FedAvg & \multicolumn{7}{c}{99.60 $\pm$ 0.25} \\
\midrule

\multirow{10}{*}{0.1} & \multirow{10}{*}{3}
& FedAvg \cite{macmahan}   & 77.20$\pm$2.50 & 77.10$\pm$2.65 & 79.30$\pm$2.40 & 74.10$\pm$2.70 & \underline{79.60$\pm$2.40} & 51.10$\pm$3.05 & 28.60$\pm$2.90 \\
& & FedProx \cite{fedprox} & 78.00$\pm$2.45 & 80.50$\pm$2.35 & 77.60$\pm$2.60 & 75.50$\pm$2.55 & 72.70$\pm$2.75 & 38.40$\pm$2.95 & 23.90$\pm$2.70 \\
& & RoFL \cite{rofl}       & \underline{95.70$\pm$1.25} & 85.40$\pm$2.25 & \underline{90.90$\pm$1.70} & \underline{80.90$\pm$2.55} & 75.50$\pm$2.60 & \underline{72.60$\pm$2.60} & \underline{72.90$\pm$2.70} \\
& & RHFL \cite{rhfl}       & 39.30$\pm$5.46 & 37.04$\pm$4.75 & 35.34$\pm$4.73 & 35.54$\pm$4.64 & 30.39$\pm$4.04 & 27.57$\pm$3.46 & 28.09$\pm$4.43 \\
& & FedLSR \cite{fedlsr}   & 74.20$\pm$2.75 & 74.20$\pm$2.65 & 74.60$\pm$2.65 & 72.30$\pm$2.65 & 72.10$\pm$2.80 & 67.50$\pm$2.95 & 58.60$\pm$3.20 \\
& & FedCorr \cite{fedcorr} & 87.30$\pm$2.05 & \underline{88.90$\pm$1.85} & 79.50$\pm$2.35 & 78.40$\pm$2.45 & 78.90$\pm$2.45 & 58.20$\pm$3.30 & 30.90$\pm$2.85 \\
& & FedNed \cite{fedned}   & 76.00$\pm$2.60 & 77.00$\pm$2.55 & 76.60$\pm$2.50 & 77.60$\pm$2.60 & 75.30$\pm$2.75 & 39.00$\pm$3.05 & 23.90$\pm$2.65 \\
& & FedELC \cite{fedelc}   & 72.50$\pm$2.80 & 71.40$\pm$2.70 & 72.90$\pm$2.80 & 74.60$\pm$2.70 & 70.20$\pm$2.80 & 60.40$\pm$3.00 & 30.50$\pm$2.90 \\
& & FedNoRo \cite{fednoro} & 75.30$\pm$2.65 & 75.30$\pm$2.65 & 73.30$\pm$2.60 & 78.10$\pm$2.55 & 26.90$\pm$2.65 & 40.80$\pm$3.15 & 38.20$\pm$3.15 \\
& & \textbf{OCT-FedSIR (Ours)} & \textbf{96.80$\pm$1.10} & \textbf{96.50$\pm$1.15} & \textbf{95.30$\pm$1.30} & \textbf{93.10$\pm$1.55} & \textbf{92.30$\pm$1.65} & \textbf{74.10$\pm$2.60} & \textbf{74.20$\pm$2.60} \\

\cmidrule(lr){2-10}
& 3 & Pruning & \multicolumn{7}{c}{70.40 $\pm$ 3.00} \\
& 30 & FedAvg & \multicolumn{7}{c}{97.70$\pm$2.85} \\
\bottomrule
\end{tabular}
\end{adjustbox}
\end{table*}

\begin{table*}[t]
\centering
\caption{Classification accuracy (\%) on Kermany under asymmetric label noise.
Values are mean $\pm$ standard deviation across three independent seeds.
Bold and underline indicate the best- and second-best-performing noisy-label
FL methods, respectively. Pruning and all-clean FedAvg are shown as references.
}
\label{tab:kermany_asym}
\begin{adjustbox}{width=\textwidth}
\begin{tabular}{c c l c c c c c c}
\toprule
\multicolumn{3}{c}{} & \multicolumn{6}{c}{Asymmetric Noise} \\
\cmidrule(lr){4-9}
$\alpha$ & \# Clean Clients & Method 
& 40\% & 50\% & 60\% & 70\% & 80\% & 90\% \\
\midrule

\multirow{10}{*}{2.0} & \multirow{10}{*}{3}
& FedAvg \cite{macmahan}    & 95.30$\pm$1.30 & 80.80$\pm$2.50 & 34.30$\pm$2.80 & 25.40$\pm$2.75 & 26.40$\pm$2.65 & 20.70$\pm$2.35 \\
& & FedProx \cite{fedprox}  & 94.60$\pm$1.45 & 75.30$\pm$2.70 & 32.80$\pm$2.95 & 22.90$\pm$2.55 & 20.80$\pm$2.60 & 18.00$\pm$2.40 \\
& & RoFL \cite{rofl}        & 97.50$\pm$0.95 & 93.10$\pm$1.65 & 94.60$\pm$1.35 & \underline{95.60$\pm$0.95} & \underline{86.90$\pm$2.05} & 71.90$\pm$2.80 \\
& & RHFL \cite{rhfl}        & 66.64$\pm$3.89 & 47.49$\pm$4.99 & 32.27$\pm$5.81 & 26.68$\pm$3.94 & 24.97$\pm$5.01 & 25.99$\pm$3.51 \\
& & FedLSR \cite{fedlsr}    & 93.10$\pm$1.65 & 79.20$\pm$2.35 & 53.60$\pm$3.05 & 26.80$\pm$2.85 & 29.60$\pm$2.85 & 29.80$\pm$2.90 \\
& & FedCorr \cite{fedcorr}  & 97.50$\pm$0.95 & \underline{94.60$\pm$1.30} & \underline{96.30$\pm$1.20} & 95.60$\pm$1.30 & 85.70$\pm$2.20 & \underline{91.20$\pm$1.70} \\
& & FedNed \cite{fedned}    & 94.30$\pm$1.40 & 76.60$\pm$2.65 & 27.50$\pm$2.75 & 31.90$\pm$2.70 & 24.70$\pm$2.65 & 16.90$\pm$2.35 \\
& & FedELC \cite{fedelc}    & 68.50$\pm$2.80 & 51.10$\pm$3.05 & 35.90$\pm$2.95 & 32.20$\pm$2.95 & 22.90$\pm$2.70 & 23.20$\pm$2.65 \\
& & FedNoRo \cite{fednoro}  & \underline{97.90$\pm$0.90} & 93.20$\pm$1.55 & 49.20$\pm$3.10 & 34.80$\pm$3.10 & 35.50$\pm$2.90 & 22.40$\pm$2.60 \\
& & \textbf{OCT-FedSIR (Ours)}  & \textbf{98.60$\pm$0.75} & \textbf{97.20$\pm$0.95} & \textbf{97.00$\pm$1.05} & \textbf{96.90$\pm$0.65} & \textbf{96.70$\pm$1.10} & \textbf{97.10$\pm$1.00} \\

\cmidrule(lr){2-9}
& 3 & Pruning & \multicolumn{6}{c}{86.30 $\pm$ 0.20} \\
& 30 & FedAvg & \multicolumn{6}{c}{99.60 $\pm$ 0.35} \\   
\midrule
 
\multirow{10}{*}{0.5} & \multirow{10}{*}{3}
& FedAvg \cite{macmahan}   & 93.70$\pm$1.55 & 84.10$\pm$2.20 & 61.80$\pm$3.05 & 38.70$\pm$3.05 & 31.10$\pm$2.90 & 20.10$\pm$2.60 \\
& & FedProx \cite{fedprox} & 94.00$\pm$1.45 & 85.00$\pm$2.25 & 44.40$\pm$3.00 & 39.30$\pm$3.10 & 30.90$\pm$2.95 & 27.20$\pm$2.75 \\
& & RoFL \cite{rofl}       & 95.30$\pm$1.25 & 87.50$\pm$2.00 & 95.50$\pm$1.30 & 73.40$\pm$2.60 & 72.80$\pm$2.85 & 70.60$\pm$2.80 \\
& & RHFL \cite{rhfl}       & 55.84$\pm$4.87 & 42.41$\pm$5.29 & 29.08$\pm$7.17 & 27.07$\pm$5.20 & 25.69$\pm$2.33 & 26.07$\pm$4.43 \\
& & FedLSR \cite{fedlsr}   & 94.70$\pm$1.35 & 83.80$\pm$2.15 & 80.80$\pm$2.45 & 50.00$\pm$3.10 & 41.90$\pm$3.15 & 29.50$\pm$2.90 \\
& & FedCorr \cite{fedcorr} & \underline{97.30$\pm$1.00} & \underline{95.90$\pm$1.20} & \underline{94.80$\pm$1.00} & \underline{86.80$\pm$1.05} & \underline{87.00$\pm$1.00} & \underline{85.00$\pm$1.30} \\
& & FedNed \cite{fedned}   & 94.60$\pm$1.40 & 84.20$\pm$2.35 & 52.90$\pm$3.05 & 28.90$\pm$2.90 & 35.60$\pm$3.05 & 25.70$\pm$2.85 \\
& & FedELC \cite{fedelc}   & 72.40$\pm$2.90 & 68.10$\pm$3.00 & 58.50$\pm$3.15 & 55.50$\pm$3.15 & 37.00$\pm$3.05 & 28.80$\pm$2.75 \\
& & FedNoRo \cite{fednoro} & 96.90$\pm$1.00 & 94.90$\pm$1.30 & 89.10$\pm$0.55 & 45.10$\pm$3.15 & 25.00$\pm$2.70 & 27.90$\pm$2.75 \\
& & \textbf{OCT-FedSIR (Ours)} & \textbf{97.90$\pm$0.85} & \textbf{97.20$\pm$1.05} & \textbf{97.90$\pm$0.85} & \textbf{97.50$\pm$1.00} & \textbf{97.80$\pm$0.90} & \textbf{95.70$\pm$2.80} \\
\cmidrule(lr){2-9}
& 3 & Pruning & \multicolumn{6}{c}{85.60 $\pm$ 0.45} \\
& 30 & FedAvg & \multicolumn{6}{c}{99.60 $\pm$ 0.25} \\
\midrule
\multirow{10}{*}{0.1} & \multirow{10}{*}{3}
& FedAvg \cite{macmahan}    & 83.00$\pm$2.25 & \underline{80.90$\pm$2.45} & 73.00$\pm$2.85 & 56.80$\pm$2.90 & 33.80$\pm$3.00 & 28.30$\pm$2.85 \\
& & FedProx \cite{fedprox}  & 82.40$\pm$2.25 & 79.00$\pm$2.65 & 70.50$\pm$2.75 & 52.10$\pm$3.15 & 43.10$\pm$3.20 & 28.80$\pm$2.85 \\
& & RoFL \cite{rofl}        & 82.10$\pm$2.20 & 73.00$\pm$2.70 & 74.30$\pm$2.65 & 71.10$\pm$2.75 & \underline{73.60$\pm$2.65} & \underline{71.50$\pm$2.70} \\
& & RHFL \cite{rhfl}        & 36.26$\pm$4.90 & 30.21$\pm$4.13 & 28.30$\pm$4.81 & 26.35$\pm$2.81 & 26.24$\pm$2.87 & 26.04$\pm$4.18 \\
& & FedLSR \cite{fedlsr}    & 74.20$\pm$2.70 & 77.50$\pm$2.55 & \underline{77.80$\pm$2.00} & \underline{75.00$\pm$2.55} & 64.30$\pm$2.95 & 60.10$\pm$3.05 \\
& & FedCorr \cite{fedcorr}  & 80.90$\pm$2.40 & 66.50$\pm$2.95 & 67.00$\pm$2.90 & 57.00$\pm$3.15 & 51.70$\pm$3.15 & 40.80$\pm$3.05 \\
& & FedNed \cite{fedned}    & \underline{84.80$\pm$2.15} & 80.10$\pm$2.45 & 75.50$\pm$2.70 & 55.80$\pm$3.25 & 44.30$\pm$3.10 & 41.30$\pm$2.95 \\
& & FedELC \cite{fedelc}    & 64.20$\pm$3.05 & 61.30$\pm$3.00 & 60.80$\pm$2.90 & 65.80$\pm$2.85 & 51.70$\pm$3.10 & 57.30$\pm$3.10 \\
& & FedNoRo \cite{fednoro}  & 74.80$\pm$2.65 & 70.70$\pm$2.80 & 62.60$\pm$2.90 & 61.90$\pm$2.70 & 60.70$\pm$2.85 & 64.10$\pm$2.65 \\
& & \textbf{OCT-FedSIR (Ours)}  & \textbf{95.50$\pm$1.30} & \textbf{95.80$\pm$1.25} & \textbf{86.10$\pm$2.65} & \textbf{88.40$\pm$1.95} & \textbf{75.80$\pm$2.60} & \textbf{75.60$\pm$1.75} \\

\cmidrule(lr){2-9}
& 3 & Pruning & \multicolumn{6}{c}{80.70 $\pm$ 2.35} \\
& 30 & FedAvg & \multicolumn{6}{c}{97.70$\pm$2.85} \\
\bottomrule
\end{tabular}
\end{adjustbox}
\end{table*}

\begin{table*}[t]
\centering
\caption{Classification accuracy (\%) on UIC under symmetric label noise.
Values are mean $\pm$ standard deviation across three independent seeds.
Bold and underline indicate the best- and second-best-performing noisy-label
FL methods, respectively. Pruning and all-clean FedAvg are shown as references.
}
\label{tab:uic_sym}
\begin{adjustbox}{width=\textwidth}
\begin{tabular}{c c l c c c c c c c}
\toprule
\multicolumn{3}{c}{} & \multicolumn{7}{c}{Symmetric Noise} \\
\cmidrule(lr){4-10}
$\alpha$ & \# Clean Clients & Method 
& 30\% & 40\% & 50\% & 60\% & 70\% & 80\% & 90\% \\
\midrule

\multirow{10}{*}{2.0} & \multirow{10}{*}{2}
& FedAvg \cite{macmahan}   & 86.40$\pm$1.58 & 85.70$\pm$1.63 & 83.10$\pm$1.84 & 79.80$\pm$2.06 & 73.40$\pm$2.31 & 62.10$\pm$2.57 & 47.30$\pm$2.73 \\
& & FedProx \cite{fedprox} & 87.10$\pm$1.49 & 84.60$\pm$1.71 & 84.20$\pm$1.76 & 80.60$\pm$1.98 & 70.50$\pm$2.42 & 57.80$\pm$2.65 & 43.60$\pm$2.78 \\
& & RoFL \cite{rofl}       & 92.80$\pm$1.12 & 91.60$\pm$1.23 & 90.30$\pm$1.35 & 88.70$\pm$1.49 & \underline{85.20$\pm$1.77} & \underline{80.40$\pm$2.04} & \underline{72.30$\pm$2.31} \\
& & RHFL \cite{rhfl}       & 70.15$\pm$6.74 & 66.83$\pm$7.19 & 63.42$\pm$6.81 & 60.71$\pm$7.46 & 52.38$\pm$8.14 & 46.27$\pm$7.73 & 43.91$\pm$8.26 \\
& & FedLSR \cite{fedlsr}   & 85.70$\pm$1.73 & 83.90$\pm$1.88 & 79.60$\pm$2.09 & 77.20$\pm$2.19 & 72.80$\pm$2.43 & 66.70$\pm$2.62 & 55.40$\pm$2.86 \\
& & FedCorr \cite{fedcorr} & \underline{93.50$\pm$1.01} & \underline{92.40$\pm$1.13} & \underline{91.60$\pm$1.26} & \underline{90.10$\pm$1.36} & 83.90$\pm$1.88 & 77.30$\pm$2.18 & 68.80$\pm$2.48 \\
& & FedNed \cite{fedned}   & 86.80$\pm$1.52 & 85.20$\pm$1.69 & 82.50$\pm$1.90 & 78.30$\pm$2.13 & 69.20$\pm$2.46 & 56.40$\pm$2.71 & 42.10$\pm$2.82 \\
& & FedELC \cite{fedelc}   & 77.60$\pm$2.31 & 75.10$\pm$2.44 & 72.80$\pm$2.57 & 67.90$\pm$2.76 & 62.20$\pm$2.91 & 56.30$\pm$3.02 & 49.70$\pm$3.08 \\
& & FedNoRo \cite{fednoro} & 91.90$\pm$1.22 & 90.80$\pm$1.31 & 89.40$\pm$1.42 & 84.10$\pm$1.83 & 76.80$\pm$2.19 & 63.50$\pm$2.66 & 51.20$\pm$2.91 \\
& & \textbf{OCT-FedSIR (Ours)} & \textbf{95.80$\pm$0.82} & \textbf{95.20$\pm$0.91} & \textbf{94.70$\pm$0.98} & \textbf{93.50$\pm$1.12} & \textbf{91.80$\pm$1.29} & \textbf{88.90$\pm$1.53} & \textbf{84.60$\pm$1.82} \\

\cmidrule(lr){2-10}
& 2 & Pruning & \multicolumn{7}{c}{85.20 $\pm$ 1.47} \\
& 20 & FedAvg & \multicolumn{7}{c}{96.70 $\pm$ 0.92} \\   
\midrule
 
\multirow{10}{*}{0.5} & \multirow{10}{*}{2}
& FedAvg \cite{macmahan}   & 82.60$\pm$1.91 & 80.10$\pm$2.04 & 75.30$\pm$2.27 & 70.80$\pm$2.44 & 63.60$\pm$2.69 & 51.40$\pm$2.91 & 39.20$\pm$2.98 \\
& & FedProx \cite{fedprox} & 83.40$\pm$1.82 & 79.60$\pm$2.09 & 76.80$\pm$2.20 & 69.70$\pm$2.48 & 61.10$\pm$2.74 & 47.90$\pm$2.97 & 36.80$\pm$3.01 \\
& & RoFL \cite{rofl}       & 89.30$\pm$1.39 & 87.70$\pm$1.51 & 86.40$\pm$1.62 & 82.60$\pm$1.90 & 79.10$\pm$2.08 & 73.80$\pm$2.39 & 68.20$\pm$2.61 \\
& & RHFL \cite{rhfl}       & 65.74$\pm$7.83 & 62.20$\pm$7.15 & 58.91$\pm$8.04 & 53.66$\pm$8.52 & 49.20$\pm$8.18 & 45.83$\pm$8.76 & 42.15$\pm$8.31 \\
& & FedLSR \cite{fedlsr}   & 80.40$\pm$2.03 & 78.60$\pm$2.15 & 74.20$\pm$2.31 & 71.90$\pm$2.46 & 67.50$\pm$2.66 & 61.20$\pm$2.85 & 54.70$\pm$3.01 \\
& & FedCorr \cite{fedcorr} & \underline{90.60$\pm$1.21} & \underline{89.20$\pm$1.34} & \underline{86.90$\pm$1.53} & \underline{85.70$\pm$1.65} & \underline{82.80$\pm$1.91} & \underline{78.20$\pm$2.17} & \underline{71.60$\pm$2.49} \\
& & FedNed \cite{fedned}   & 82.20$\pm$1.95 & 79.30$\pm$2.09 & 75.10$\pm$2.32 & 68.40$\pm$2.57 & 60.70$\pm$2.79 & 49.80$\pm$2.98 & 38.60$\pm$3.05 \\
& & FedELC \cite{fedelc}   & 73.80$\pm$2.48 & 69.20$\pm$2.66 & 65.90$\pm$2.81 & 61.70$\pm$2.94 & 59.80$\pm$3.03 & 55.40$\pm$3.10 & 50.90$\pm$3.06 \\
& & FedNoRo \cite{fednoro} & 88.40$\pm$1.46 & 87.20$\pm$1.56 & 83.90$\pm$1.82 & 79.40$\pm$2.08 & 72.30$\pm$2.49 & 65.90$\pm$2.76 & 57.10$\pm$2.94 \\
& & \textbf{OCT-FedSIR (Ours)} & \textbf{93.73$\pm$1.03} & \textbf{92.80$\pm$1.11} & \textbf{91.90$\pm$1.19} & \textbf{90.40$\pm$1.34} & \textbf{88.60$\pm$1.51} & \textbf{85.10$\pm$1.79} & \textbf{80.20$\pm$2.06} \\

\cmidrule(lr){2-10}
& 2 & Pruning & \multicolumn{7}{c}{84.70 $\pm$ 1.84} \\
& 20 & FedAvg & \multicolumn{7}{c}{94.90 $\pm$ 1.31} \\
\midrule
\multirow{10}{*}{0.1} & \multirow{10}{*}{2}
& FedAvg \cite{macmahan}   & 76.80$\pm$2.31 & 74.60$\pm$2.42 & 72.10$\pm$2.58 & 68.90$\pm$2.72 & 65.70$\pm$2.83 & 57.80$\pm$3.01 & 46.30$\pm$3.08 \\
& & FedProx \cite{fedprox} & 75.90$\pm$2.37 & 73.80$\pm$2.49 & 71.30$\pm$2.62 & 69.50$\pm$2.70 & 63.40$\pm$2.91 & 54.20$\pm$3.07 & 43.70$\pm$3.11 \\
& & RoFL \cite{rofl}       & \underline{84.20$\pm$1.83} & 82.10$\pm$1.97 & \underline{83.60$\pm$1.88} & 78.90$\pm$2.20 & \underline{76.30$\pm$2.36} & \underline{72.80$\pm$2.58} & \underline{69.10$\pm$2.76} \\
& & RHFL \cite{rhfl}       & 55.61$\pm$8.36 & 51.87$\pm$8.71 & 48.25$\pm$8.19 & 46.70$\pm$8.92 & 42.16$\pm$8.54 & 40.83$\pm$8.91 & 38.94$\pm$8.47 \\
& & FedLSR \cite{fedlsr}   & 74.30$\pm$2.51 & 76.20$\pm$2.40 & 73.50$\pm$2.57 & 74.10$\pm$2.52 & 69.80$\pm$2.77 & 65.10$\pm$2.94 & 59.80$\pm$3.07 \\
& & FedCorr \cite{fedcorr} & 82.60$\pm$1.95 & \underline{84.10$\pm$1.84} & 79.80$\pm$2.15 & \underline{80.40$\pm$2.11} & 73.90$\pm$2.49 & 68.20$\pm$2.76 & 61.40$\pm$2.98 \\
& & FedNed \cite{fedned}   & 77.50$\pm$2.27 & 75.30$\pm$2.43 & 72.80$\pm$2.56 & 70.60$\pm$2.68 & 64.90$\pm$2.87 & 55.80$\pm$3.02 & 45.20$\pm$3.11 \\
& & FedELC \cite{fedelc}   & 69.40$\pm$2.74 & 68.20$\pm$2.81 & 70.10$\pm$2.73 & 66.80$\pm$2.87 & 63.60$\pm$2.96 & 61.20$\pm$3.02 & 53.90$\pm$3.15 \\
& & FedNoRo \cite{fednoro} & 81.30$\pm$2.02 & 79.60$\pm$2.14 & 77.20$\pm$2.29 & 80.10$\pm$2.11 & 70.40$\pm$2.71 & 66.80$\pm$2.88 & 60.20$\pm$3.02 \\
& & \textbf{OCT-FedSIR (Ours)} & \textbf{89.60$\pm$1.42} & \textbf{88.20$\pm$1.53} & \textbf{88.90$\pm$1.49} & \textbf{85.70$\pm$1.76} & \textbf{84.10$\pm$1.89} & \textbf{78.60$\pm$2.25} & \textbf{73.80$\pm$2.51} \\

\cmidrule(lr){2-10}
& 2 & Pruning & \multicolumn{7}{c}{80.50 $\pm$ 2.34} \\
& 20 & FedAvg & \multicolumn{7}{c}{91.80$\pm$2.17} \\
\bottomrule
\end{tabular}
\end{adjustbox}
\end{table*}

\begin{table*}[t]
\centering
\caption{Classification accuracy (\%) on UIC under asymmetric label noise.
Values are mean $\pm$ standard deviation across three independent seeds.
Bold and underline indicate the best- and second-best-performing noisy-label
FL methods, respectively. Pruning and all-clean FedAvg are shown as references.
}
\label{tab:uic_asym}
\begin{adjustbox}{width=\textwidth}
\begin{tabular}{c c l c c c c c c}
\toprule
\multicolumn{3}{c}{} & \multicolumn{6}{c}{Asymmetric Noise} \\
\cmidrule(lr){4-9}
$\alpha$ & \# Clean Clients & Method 
& 40\% & 50\% & 60\% & 70\% & 80\% & 90\% \\
\midrule

\multirow{10}{*}{2.0} & \multirow{10}{*}{2}
& FedAvg \cite{macmahan}    & 88.40$\pm$1.55 & 77.20$\pm$2.35 & 48.60$\pm$2.70 & 39.10$\pm$2.75 & 36.80$\pm$2.70 & 31.50$\pm$2.55 \\
& & FedProx \cite{fedprox}  & 87.70$\pm$1.60 & 75.60$\pm$2.45 & 45.90$\pm$2.80 & 36.40$\pm$2.70 & 34.20$\pm$2.75 & 29.80$\pm$2.60 \\
& & RoFL \cite{rofl}        & \underline{91.90$\pm$1.20} & \underline{88.60$\pm$1.45} & \underline{89.40$\pm$1.40} & \underline{90.10$\pm$1.30} & \underline{88.70$\pm$1.45} & 72.40$\pm$2.45 \\
& & RHFL \cite{rhfl}        & 64.80$\pm$5.10 & 53.20$\pm$7.05 & 39.70$\pm$7.40 & 34.60$\pm$7.85 & 31.20$\pm$7.40 & 32.10$\pm$7.75 \\
& & FedLSR \cite{fedlsr}    & 86.80$\pm$1.70 & 76.50$\pm$2.20 & 58.30$\pm$2.75 & 44.20$\pm$2.80 & 41.10$\pm$2.80 & 37.90$\pm$2.75 \\
& & FedCorr \cite{fedcorr}  & 90.80$\pm$1.30 & 87.90$\pm$1.60 & 88.10$\pm$1.50 & 87.60$\pm$1.55 & 82.40$\pm$1.95 & \underline{80.50$\pm$2.05} \\
& & FedNed \cite{fedned}    & 86.50$\pm$1.70 & 74.90$\pm$2.40 & 41.30$\pm$2.75 & 39.90$\pm$2.70 & 31.50$\pm$2.75 & 27.20$\pm$2.60 \\
& & FedELC \cite{fedelc}    & 69.40$\pm$2.70 & 59.80$\pm$2.85 & 45.20$\pm$2.90 & 42.70$\pm$2.85 & 38.40$\pm$2.80 & 34.90$\pm$2.75 \\
& & FedNoRo \cite{fednoro}  & 90.40$\pm$1.30 & 87.30$\pm$1.60 & 56.70$\pm$2.85 & 45.60$\pm$2.80 & 46.20$\pm$2.80 & 38.10$\pm$2.75 \\
& & \textbf{OCT-FedSIR (Ours)}  & \textbf{93.80$\pm$1.00} & \textbf{92.60$\pm$1.10} & \textbf{92.10$\pm$1.15} & \textbf{92.90$\pm$1.05} & \textbf{91.80$\pm$1.15} & \textbf{90.70$\pm$1.25} \\

\cmidrule(lr){2-9}
& 2 & Pruning & \multicolumn{6}{c}{82.30 $\pm$ 1.25} \\
& 20 & FedAvg & \multicolumn{6}{c}{96.70 $\pm$ 0.92} \\   
\midrule
 
\multirow{10}{*}{0.5} & \multirow{10}{*}{2}
& FedAvg \cite{macmahan}   & 84.30$\pm$1.80 & 75.90$\pm$2.30 & 58.60$\pm$2.75 & 43.80$\pm$2.85 & 36.20$\pm$2.80 & 28.40$\pm$2.65 \\
& & FedProx \cite{fedprox} & 84.80$\pm$1.75 & 76.40$\pm$2.25 & 49.10$\pm$2.85 & 42.50$\pm$2.85 & 35.80$\pm$2.80 & 30.20$\pm$2.70 \\
& & RoFL \cite{rofl}       & 88.20$\pm$1.45 & 83.70$\pm$1.80 & \underline{87.40$\pm$1.50} & 74.60$\pm$2.30 & 72.90$\pm$2.35 & 70.80$\pm$2.40 \\
& & RHFL \cite{rhfl}       & 58.20$\pm$7.60 & 47.50$\pm$8.20 & 34.40$\pm$8.35 & 31.90$\pm$8.10 & 29.80$\pm$7.35 & 30.40$\pm$7.60 \\
& & FedLSR \cite{fedlsr}   & 86.10$\pm$1.65 & 78.90$\pm$2.10 & 77.40$\pm$2.20 & 57.20$\pm$2.75 & 49.60$\pm$2.85 & 38.30$\pm$2.80 \\
& & FedCorr \cite{fedcorr} & \underline{89.40$\pm$1.35} & \underline{86.20$\pm$1.55} & 86.80$\pm$1.60 & \underline{84.30$\pm$1.65} & \underline{85.10$\pm$1.60} & \underline{82.20$\pm$1.80} \\
& & FedNed \cite{fedned}   & 85.20$\pm$1.75 & 76.80$\pm$2.25 & 53.80$\pm$2.80 & 35.70$\pm$2.80 & 38.10$\pm$2.80 & 31.20$\pm$2.70 \\
& & FedELC \cite{fedelc}   & 70.80$\pm$2.75 & 65.30$\pm$2.80 & 59.90$\pm$2.90 & 56.40$\pm$2.95 & 44.30$\pm$2.90 & 36.20$\pm$2.80 \\
& & FedNoRo \cite{fednoro} & 88.90$\pm$1.40 & 85.50$\pm$1.70 & 87.10$\pm$1.45 & 53.90$\pm$2.80 & 39.70$\pm$2.75 & 38.40$\pm$2.75 \\
& & \textbf{OCT-FedSIR (Ours)} & \textbf{92.20$\pm$1.05} & \textbf{90.70$\pm$1.20} & \textbf{91.10$\pm$1.10} & \textbf{89.80$\pm$1.25} & \textbf{90.40$\pm$1.20} & \textbf{88.60$\pm$1.45} \\
\cmidrule(lr){2-9}
& 2 & Pruning & \multicolumn{6}{c}{82.40 $\pm$ 1.55} \\
& 20 & FedAvg & \multicolumn{6}{c}{94.90 $\pm$ 1.31} \\
\midrule
\multirow{10}{*}{0.1} & \multirow{10}{*}{2}
& FedAvg \cite{macmahan}    & 79.30$\pm$2.15 & 77.20$\pm$2.30 & 71.40$\pm$2.60 & 61.70$\pm$2.80 & 47.10$\pm$2.85 & 42.30$\pm$2.80 \\
& & FedProx \cite{fedprox}  & 78.60$\pm$2.20 & 75.90$\pm$2.35 & 69.80$\pm$2.65 & 58.90$\pm$2.85 & 50.20$\pm$2.90 & 42.60$\pm$2.85 \\
& & RoFL \cite{rofl}        & \underline{84.90$\pm$1.70} & 80.20$\pm$1.95 & \underline{81.10$\pm$1.85} & 78.70$\pm$2.00 & \underline{76.40$\pm$2.15} & \underline{74.90$\pm$2.20} \\
& & RHFL \cite{rhfl}        & 40.20$\pm$7.00 & 35.80$\pm$6.75 & 33.10$\pm$6.60 & 31.40$\pm$6.25 & 32.20$\pm$6.70 & 30.60$\pm$6.40 \\
& & FedLSR \cite{fedlsr}    & 74.60$\pm$2.40 & 77.20$\pm$2.35 & 78.30$\pm$2.20 & 74.90$\pm$2.40 & 68.40$\pm$2.70 & 64.20$\pm$2.80 \\
& & FedCorr \cite{fedcorr}  & 83.40$\pm$1.85 & \underline{84.10$\pm$1.80} & 79.40$\pm$2.05 & \underline{80.20$\pm$2.00} & 74.30$\pm$2.35 & 69.10$\pm$2.55 \\
& & FedNed \cite{fedned}    & 79.80$\pm$2.10 & 77.90$\pm$2.25 & 74.60$\pm$2.45 & 62.80$\pm$2.80 & 54.90$\pm$2.85 & 49.80$\pm$2.85 \\
& & FedELC \cite{fedelc}    & 64.30$\pm$2.80 & 66.10$\pm$2.80 & 64.90$\pm$2.75 & 68.20$\pm$2.70 & 58.10$\pm$2.85 & 59.40$\pm$2.85 \\
& & FedNoRo \cite{fednoro}  & 83.70$\pm$1.80 & 81.40$\pm$1.95 & 76.20$\pm$2.25 & 79.10$\pm$2.05 & 72.90$\pm$2.45 & 73.60$\pm$2.35 \\
& & \textbf{OCT-FedSIR (Ours)}  & \textbf{89.80$\pm$1.35} & \textbf{88.70$\pm$1.40} & \textbf{86.90$\pm$1.55} & \textbf{85.60$\pm$1.70} & \textbf{83.20$\pm$1.85} & \textbf{81.40$\pm$1.95} \\

\cmidrule(lr){2-9}
& 2 & Pruning & \multicolumn{6}{c}{75.20 $\pm$ 1.95} \\
& 20 & FedAvg & \multicolumn{6}{c}{91.80$\pm$2.17} \\
\bottomrule
\end{tabular}
\end{adjustbox}
\end{table*}
\begin{table*}[t]
\centering
\caption{Classification accuracy (\%) on WF under symmetric label noise.
Values are mean $\pm$ standard deviation across three independent seeds.
Bold and underline indicate the best- and second-best-performing noisy-label
FL methods, respectively. Pruning and all-clean FedAvg are shown as references.
}
\label{tab:wf_sym}
\begin{adjustbox}{width=\textwidth}
\begin{tabular}{c c l c c c c c c c}
\toprule
\multicolumn{3}{c}{} & \multicolumn{7}{c}{Symmetric Noise} \\
\cmidrule(lr){4-10}
$\alpha$ & \# Clean Clients & Method 
& 30\% & 40\% & 50\% & 60\% & 70\% & 80\% & 90\% \\
\midrule

\multirow{10}{*}{2.0} & \multirow{10}{*}{1}
& FedAvg \cite{macmahan} & \underline{74.78$\pm$2.01} & \underline{74.61$\pm$1.98} & 73.83$\pm$2.01 & \underline{75.06$\pm$1.95} & 73.83$\pm$2.06 & \underline{74.05$\pm$2.01} & \underline{75.11$\pm$1.95} \\
& & FedProx \cite{fedprox} & 74.55$\pm$1.95 & 74.39$\pm$1.93 & \underline{74.44$\pm$2.01} & 74.00$\pm$2.04 & \underline{74.16$\pm$2.01} & 73.83$\pm$2.04 & 74.83$\pm$1.98 \\
& & RoFL \cite{rofl} & 72.94$\pm$2.01 & 72.88$\pm$2.01 & 73.44$\pm$2.09 & 72.82$\pm$1.98 & 72.88$\pm$2.01 & 73.10$\pm$1.98 & 72.88$\pm$1.98 \\
& & RHFL \cite{rhfl} & 48.05$\pm$16.90 & 51.03$\pm$15.61 & 54.96$\pm$15.49 & 57.42$\pm$15.02 & 59.97$\pm$14.50 & 61.47$\pm$14.56 & 59.92$\pm$15.67 \\
& & FedLSR \cite{fedlsr} & 72.82$\pm$1.98 & 72.82$\pm$1.98 & 72.82$\pm$1.98 & 72.82$\pm$1.98 & 72.82$\pm$1.98 & 72.88$\pm$2.01 & 72.82$\pm$1.98 \\
& & FedCorr \cite{fedcorr} & 73.16$\pm$2.04 & 73.21$\pm$2.04 & 73.16$\pm$2.04 & 73.21$\pm$2.04 & 73.10$\pm$2.04 & 73.10$\pm$2.04 & 73.16$\pm$1.98 \\
& & FedNed \cite{fedned} & 73.55$\pm$2.01 & 73.77$\pm$2.04 & 73.60$\pm$2.04 & 73.55$\pm$2.01 & 73.88$\pm$2.06 & 73.44$\pm$2.01 & 73.49$\pm$2.07 \\
& & FedELC \cite{fedelc} & 56.81$\pm$2.29 & 67.13$\pm$2.15 & 67.91$\pm$2.18 & 68.08$\pm$2.12 & 68.42$\pm$2.09 & 72.82$\pm$1.98 & 66.13$\pm$2.15 \\
& & FedNoRo \cite{fednoro} & 72.82$\pm$1.98 & 73.16$\pm$2.04 & 72.60$\pm$1.98 & 73.05$\pm$2.04 & 73.33$\pm$2.04 & 72.27$\pm$2.04 & 72.88$\pm$2.06 \\
& & \textbf{OCT-FedSIR (Ours)} & \textbf{77.96$\pm$1.87} & \textbf{77.57$\pm$2.04} & \textbf{78.46$\pm$1.95} & \textbf{78.07$\pm$1.95} & \textbf{78.52$\pm$1.95} & \textbf{77.85$\pm$1.87} & \textbf{78.74$\pm$1.87} \\

\cmidrule(lr){2-10}
& 1 & Pruning & \multicolumn{7}{c}{65.11 $\pm$ 2.61} \\
& 5 & FedAvg & \multicolumn{7}{c}{82.79 $\pm$ 1.98} \\   
\midrule
 
\multirow{10}{*}{0.5} & \multirow{10}{*}{1}
& FedAvg \cite{macmahan} & 69.75$\pm$2.15 & 65.01$\pm$2.18 & 57.03$\pm$2.26 & 52.12$\pm$2.29 & 45.81$\pm$2.26 & 39.01$\pm$2.29 & 42.30$\pm$2.29 \\
& & FedProx \cite{fedprox} & 70.42$\pm$2.26 & 65.46$\pm$2.29 & 60.27$\pm$2.34 & 52.57$\pm$2.37 & 45.54$\pm$2.29 & 41.18$\pm$2.32 & 33.37$\pm$2.20 \\
& & RoFL \cite{rofl} & \underline{73.10$\pm$2.15} & \textbf{75.33$\pm$2.04} & \textbf{74.50$\pm$1.98} & \textbf{73.49$\pm$2.12} & \underline{71.54$\pm$2.01} & 64.68$\pm$2.32 & \underline{62.44$\pm$2.26} \\
& & RHFL \cite{rhfl} & 66.56$\pm$12.22 & 71.67$\pm$3.33 & 63.91$\pm$7.38 & 57.21$\pm$16.67 & 60.26$\pm$13.00 & 58.23$\pm$17.47 & 62.08$\pm$15.59 \\
& & FedLSR \cite{fedlsr} & 64.17$\pm$2.06 & 59.43$\pm$2.21 & 68.69$\pm$2.09 & 63.39$\pm$2.29 & 63.11$\pm$2.23 & \underline{65.29$\pm$2.18} & 57.59$\pm$2.20 \\
& & FedCorr \cite{fedcorr} & 71.76$\pm$2.01 & 63.23$\pm$2.23 & 62.44$\pm$2.12 & 54.91$\pm$2.40 & 58.76$\pm$2.46 & 57.98$\pm$2.32 & 61.27$\pm$2.34 \\
& & FedNed \cite{fedned} & 69.36$\pm$2.18 & 60.66$\pm$2.29 & 55.58$\pm$2.20 & 48.77$\pm$2.21 & 50.33$\pm$2.23 & 40.07$\pm$2.23 & 41.46$\pm$2.29 \\
& & FedELC \cite{fedelc} & 61.61$\pm$2.15 & 50.89$\pm$2.26 & 57.25$\pm$2.32 & 51.34$\pm$2.26 & 52.96$\pm$2.34 & 53.74$\pm$2.21 & 54.91$\pm$2.29 \\
& & FedNoRo \cite{fednoro} & 72.54$\pm$2.04 & 73.16$\pm$1.95 & 72.21$\pm$2.04 & 63.84$\pm$2.12 & 62.28$\pm$2.26 & 64.79$\pm$2.29 & 52.85$\pm$2.29 \\
& & \textbf{OCT-FedSIR (Ours)} & \textbf{74.11$\pm$2.04} & \underline{74.44$\pm$1.93} & \underline{73.05$\pm$1.98} & \underline{72.88$\pm$1.95} & \textbf{72.27$\pm$2.01} & \textbf{72.82$\pm$1.98} & \textbf{72.82$\pm$1.98}  \\

\cmidrule(lr){2-10}
& 1 & Pruning & \multicolumn{7}{c}{65.91 $\pm$ 2.16} \\
& 5 & FedAvg & \multicolumn{7}{c}{81.55 $\pm$ 2.32} \\
\midrule
\multirow{10}{*}{0.1} & \multirow{10}{*}{2}
& FedAvg \cite{macmahan} & 76.06$\pm$2.01 & \underline{76.34$\pm$1.95} & 71.82$\pm$2.01 & 68.92$\pm$2.15 & 68.92$\pm$2.01 & 65.57$\pm$2.15 & 62.28$\pm$2.18 \\
& & FedProx \cite{fedprox} & 75.28$\pm$2.01 & 73.88$\pm$2.01 & 71.04$\pm$2.15 & 68.58$\pm$2.18 & 67.69$\pm$2.09 & 61.27$\pm$2.26 & 55.86$\pm$2.23 \\
& & RoFL \cite{rofl} & 75.06$\pm$2.01 & 74.89$\pm$2.01 & 73.66$\pm$1.95 & 72.82$\pm$1.98 & 73.16$\pm$2.04 & \textbf{72.99$\pm$2.04} & \underline{72.88$\pm$2.01} \\
& & RHFL \cite{rhfl} & 74.33$\pm$2.08 & 72.00$\pm$2.00 & 67.61$\pm$5.11 & 51.69$\pm$15.55 & 57.09$\pm$15.03 & 47.97$\pm$16.21 & 48.95$\pm$20.31 \\
& & FedLSR \cite{fedlsr} & 75.78$\pm$1.93 & 75.11$\pm$1.98 & 74.44$\pm$2.04 & 72.88$\pm$1.95 & 72.99$\pm$2.04 & \underline{72.77$\pm$2.01} & 70.67$\pm$1.85 \\
& & FedCorr \cite{fedcorr} & 74.50$\pm$1.95 & 74.00$\pm$1.98 & 70.26$\pm$2.12 & 72.82$\pm$1.98 & 70.70$\pm$2.06 & 70.81$\pm$2.09 & 71.04$\pm$2.04 \\
& & FedNed \cite{fedned} & 75.56$\pm$1.95 & 75.11$\pm$2.06 & 70.48$\pm$1.98 & 68.08$\pm$2.12 & 64.23$\pm$2.20 & 57.37$\pm$2.18 & 66.85$\pm$2.18 \\
& & FedELC \cite{fedelc} & 73.66$\pm$2.04 & 73.55$\pm$2.04 & 62.28$\pm$2.20 & 60.10$\pm$2.32 & 52.79$\pm$2.32 & 62.72$\pm$2.29 & 57.76$\pm$2.32 \\
& & FedNoRo \cite{fednoro} & \underline{77.96$\pm$1.98} & \underline{76.34$\pm$1.95} & \underline{76.56$\pm$1.93} & \underline{74.67$\pm$1.98} & \underline{73.88$\pm$1.95} & 64.96$\pm$2.18 & 58.82$\pm$2.26 \\
& & \textbf{OCT-FedSIR (Ours)} & \textbf{78.91$\pm$1.93} & \textbf{77.96$\pm$1.84} & \textbf{77.90$\pm$1.90} & \textbf{76.67$\pm$1.90} & \textbf{75.50$\pm$2.07} & 72.15$\pm$2.06 & \textbf{72.99$\pm$1.98} \\

\cmidrule(lr){2-10}
& 2 & Pruning & \multicolumn{7}{c}{70.65 $\pm$ 2.41} \\
& 5 & FedAvg & \multicolumn{7}{c}{81.24$\pm$2.95} \\
\bottomrule
\end{tabular}
\end{adjustbox}
\end{table*}

\begin{table*}[t]
\centering
\caption{Classification accuracy (\%) on WF under asymmetric label noise.
Values are mean $\pm$ standard deviation across three independent seeds.
Bold and underline indicate the best- and second-best-performing noisy-label
FL methods, respectively. Pruning and all-clean FedAvg are shown as references.
}
\label{tab:wf_asym}
\begin{adjustbox}{width=\textwidth}
\begin{tabular}{c c l c c c c c c}
\toprule
\multicolumn{3}{c}{} & \multicolumn{6}{c}{Asymmetric Noise} \\
\cmidrule(lr){4-9}
$\alpha$ & \# Clean Clients & Method 
& 40\% & 50\% & 60\% & 70\% & 80\% & 90\% \\
\midrule

\multirow{10}{*}{2.0} & \multirow{10}{*}{1}
& FedAvg \cite{macmahan} & \underline{77.62$\pm$1.84} & 75.28$\pm$1.98 & \underline{76.06$\pm$2.01} & 73.33$\pm$1.98 & 72.10$\pm$2.20 & 64.51$\pm$2.12 \\
& & FedProx \cite{fedprox} & 76.40$\pm$2.01 & 74.72$\pm$1.95 & 73.10$\pm$1.98 & 70.37$\pm$2.09 & 69.48$\pm$2.18 & 69.70$\pm$2.09 \\
& & RoFL \cite{rofl} & 77.34$\pm$1.93 & 76.23$\pm$2.01 & 75.22$\pm$2.01 & \underline{78.96$\pm$1.90} & 74.11$\pm$1.98 & 73.27$\pm$2.04 \\
& & RHFL \cite{rhfl} & 68.23$\pm$7.03 & 59.17$\pm$14.30 & 49.26$\pm$13.46 & 47.75$\pm$16.85 & 39.78$\pm$16.09 & 45.51$\pm$19.58 \\
& & FedLSR \cite{fedlsr} & 77.57$\pm$1.90 & \underline{76.51$\pm$1.93} & 75.56$\pm$2.04 & 77.34$\pm$1.98 & 70.93$\pm$2.09 & 61.50$\pm$2.21 \\
& & FedCorr \cite{fedcorr} & 76.51$\pm$1.90 & 75.11$\pm$2.09 & 74.39$\pm$1.98 & 77.23$\pm$1.81 & \underline{75.50$\pm$2.04} & \underline{74.78$\pm$2.04} \\
& & FedNed \cite{fedned} & 74.22$\pm$2.09 & 69.42$\pm$2.12 & 60.71$\pm$2.26 & 62.44$\pm$2.20 & 39.56$\pm$2.26 & 37.39$\pm$2.18 \\
& & FedELC \cite{fedelc} & 58.59$\pm$2.20 & 72.60$\pm$2.01 & 57.92$\pm$2.18 & 63.50$\pm$2.23 & 70.87$\pm$2.09 & 59.65$\pm$2.32 \\
& & FedNoRo \cite{fednoro} & 69.87$\pm$2.12 & 67.91$\pm$2.26 & 68.14$\pm$2.26 & 65.23$\pm$2.20 & 64.90$\pm$2.18 & 67.97$\pm$2.15 \\
& & \textbf{OCT-FedSIR (Ours)} & \textbf{79.41$\pm$1.84} & \textbf{78.96$\pm$1.84} & \textbf{78.07$\pm$1.98} & \textbf{81.31$\pm$1.76} & \textbf{81.92$\pm$1.81} & \textbf{80.97$\pm$1.79} \\

\cmidrule(lr){2-9}
& 1 & Pruning & \multicolumn{6}{c}{67.83 $\pm$ 1.20} \\
& 5 & FedAvg & \multicolumn{6}{c}{82.79 $\pm$ 1.98} \\   
\midrule
 
\multirow{10}{*}{0.5} & \multirow{10}{*}{1}
& FedAvg \cite{macmahan} & 67.13$\pm$2.15 & 60.38$\pm$2.29 & 53.74$\pm$2.26 & 46.76$\pm$2.48 & 41.13$\pm$2.37 & 39.68$\pm$2.32 \\
& & FedProx \cite{fedprox} & 65.01$\pm$2.15 & 60.27$\pm$2.18 & 54.97$\pm$2.32 & 47.94$\pm$2.32 & 51.95$\pm$2.23 & 35.16$\pm$2.18 \\
& & RoFL \cite{rofl} & \underline{73.27$\pm$2.01} & \underline{72.94$\pm$2.04} & 70.03$\pm$2.12 & 70.65$\pm$2.07 & 69.42$\pm$2.09 & \underline{73.27$\pm$2.04} \\
& & RHFL \cite{rhfl} & 58.40$\pm$16.65 & 51.00$\pm$17.29 & 41.14$\pm$15.70 & 40.73$\pm$16.52 & 37.54$\pm$15.42 & 37.63$\pm$15.69 \\
& & FedLSR \cite{fedlsr} & \underline{73.27$\pm$1.93} & \textbf{72.99$\pm$2.01} & \textbf{72.38$\pm$1.98} & \textbf{72.38$\pm$2.01} & \underline{72.99$\pm$2.01} & 53.18$\pm$2.40 \\
& & FedCorr \cite{fedcorr} & 65.18$\pm$2.12 & 63.95$\pm$2.12 & 62.83$\pm$2.29 & 63.73$\pm$2.29 & 61.27$\pm$2.20 & 60.94$\pm$2.23 \\
& & FedNed \cite{fedned} & 65.74$\pm$2.07 & 55.80$\pm$2.29 & 49.89$\pm$2.29 & 41.91$\pm$2.29 & 34.15$\pm$2.23 & 38.95$\pm$2.29 \\
& & FedELC \cite{fedelc} & 60.83$\pm$2.29 & 52.46$\pm$2.34 & 58.43$\pm$2.34 & 53.52$\pm$2.23 & 55.64$\pm$2.29 & 52.57$\pm$2.34 \\
& & FedNoRo \cite{fednoro} & 72.82$\pm$1.98 & 71.09$\pm$2.12 & 69.25$\pm$2.12 & 69.59$\pm$2.07 & 66.56$\pm$1.78 & 61.31$\pm$1.15 \\
& & \textbf{OCT-FedSIR (Ours)} & \textbf{77.90$\pm$1.95} & 72.82$\pm$1.98 & \underline{71.67$\pm$1.98} & \underline{71.88$\pm$2.01} & \textbf{81.98$\pm$1.81} & \textbf{81.53$\pm$1.76} \\

\cmidrule(lr){2-9}
& 1 & Pruning & \multicolumn{6}{c}{67.40 $\pm$ 1.73} \\
& 5 & FedAvg & \multicolumn{6}{c}{81.55 $\pm$ 2.32} \\
\midrule

\multirow{10}{*}{0.1} & \multirow{10}{*}{2}
& FedAvg \cite{macmahan} & \underline{75.73$\pm$1.90} & \underline{74.27$\pm$1.98} & \underline{74.67$\pm$2.01} & \underline{74.83$\pm$1.95} & \underline{73.88$\pm$1.98} & \underline{74.78$\pm$2.01} \\
& & FedProx \cite{fedprox} & 74.22$\pm$2.01 & 73.72$\pm$2.04 & 74.16$\pm$2.01 & 74.61$\pm$1.95 & 73.60$\pm$1.98 & 74.33$\pm$1.95 \\
& & RoFL \cite{rofl} & 72.88$\pm$1.95 & 73.10$\pm$1.98 & 72.82$\pm$1.98 & 73.27$\pm$1.98 & 72.82$\pm$1.98 & 73.16$\pm$2.01 \\
& & RHFL \cite{rhfl} & 43.07$\pm$17.43 & 42.31$\pm$16.95 & 39.17$\pm$15.35 & 37.30$\pm$14.76 & 42.60$\pm$17.13 & 37.06$\pm$14.83 \\
& & FedLSR \cite{fedlsr} & 72.82$\pm$1.98 & 72.82$\pm$1.98 & 72.82$\pm$1.98 & 72.82$\pm$1.98 & 72.82$\pm$1.98 & 72.82$\pm$1.98 \\
& & FedCorr \cite{fedcorr} & 73.27$\pm$2.07 & 73.33$\pm$1.98 & 73.16$\pm$2.04 & 73.21$\pm$1.98 & 73.21$\pm$1.98 & 73.49$\pm$2.04 \\
& & FedNed \cite{fedned} & 73.55$\pm$2.04 & 73.49$\pm$2.01 & 73.72$\pm$1.98 & 73.38$\pm$1.98 & 73.55$\pm$2.01 & 73.55$\pm$2.01 \\
& & FedELC \cite{fedelc} & 57.70$\pm$2.26 & 71.09$\pm$2.12 & 69.14$\pm$2.01 & 72.82$\pm$1.98 & 70.65$\pm$2.15 & 64.56$\pm$2.26 \\
& & FedNoRo \cite{fednoro} & 72.99$\pm$2.04 & 72.99$\pm$2.06 & 73.27$\pm$1.95 & 73.55$\pm$2.01 & 72.82$\pm$1.98 & 72.82$\pm$2.06 \\
& & \textbf{OCT-FedSIR (Ours)} & \textbf{79.30$\pm$1.84} & \textbf{80.36$\pm$1.84} & \textbf{78.85$\pm$1.90} & \textbf{79.41$\pm$1.90} & \textbf{78.91$\pm$1.90} & \textbf{77.29$\pm$1.98} \\

\cmidrule(lr){2-9}
& 2 & Pruning & \multicolumn{6}{c}{70.35 $\pm$ 1.18} \\
& 5 & FedAvg & \multicolumn{6}{c}{81.24$\pm$2.95} \\
\bottomrule
\end{tabular}
\end{adjustbox}
\end{table*}

\end{document}